%% file: main.tex
\documentclass[runningheads]{llncs}
\usepackage{graphicx}

\def\MYTITLE{Secrets of Event-Based Optical Flow}

\usepackage{tikz}
\usepackage{comment}
\usepackage{amsmath,amssymb,amsfonts}
\usepackage{color}
\usepackage[breaklinks=true,bookmarks=false,hidelinks]{hyperref} %
\hypersetup{
  pdftitle={\MYTITLE},
  pdfsubject={Computer Vision, Robotics, Deep Learning},
  pdfauthor={Shintaro Shiba, Yoshimitsu Aoki, Guillermo Gallego},
  pdfkeywords={Event Cameras, Optical flow, Asynchronous Sensor, High Temporal Resolution}
}
\usepackage{orcidlink}
\usepackage{booktabs}
\usepackage{tabularx}
\usepackage{arydshln}

\usepackage{siunitx}
\usepackage{etoolbox}
\robustify\bfseries

\usepackage{makecell}
\usepackage{adjustbox} %
\usepackage{array}
\usepackage{multirow}
\usepackage{rotating}
\usepackage[english]{babel}

\usepackage{cite}
\usepackage{url}
\input{chapters/macros}

\title{\MYTITLE}
\usepackage[absolute]{textpos}

\begin{document}

\definecolor{somegray}{gray}{0.6}
\newcommand{\darkgrayed}[1]{\textcolor{somegray}{#1}}
\begin{textblock}{8}(4, 0.8)
\begin{center}
\darkgrayed{This paper has been accepted for publication at the \\
European Conference on Computer Vision (ECCV), Tel Aviv, 2022.}
\end{center}
\end{textblock}

\pagestyle{headings}
\mainmatter

\titlerunning{\MYTITLE}  %
\author{Shintaro Shiba\inst{1,2}\orcidlink{0000-0001-6053-2285} 
\and Yoshimitsu Aoki\inst{1} \and
Guillermo Gallego\inst{2,3}\orcidlink{0000-0002-2672-9241}
}
\authorrunning{S. Shiba et al.}

\institute{%
Department of Electronics and Electrical Engineering, Faculty of Science and Technology, Keio University, 
Kanagawa, Japan.  
\email{sshiba@keio.jp} \and
Department of EECS, Technische Universität Berlin, Berlin, Germany \and
Einstein Center Digital Future and SCIoI
Excellence Cluster, Berlin, Germany %
}
\maketitle

\input{chapters/00_abstract}

\input{chapters/01_intro}
\input{chapters/02_related}
\input{chapters/03_method}
\input{chapters/04_experiments}

\input{chapters/05_limitations}

\input{chapters/06_conclusion}

\input{chapters/06_ack}

\clearpage \input{chapters/07_suppl_mat}

\clearpage
\bibliographystyle{splncs04}
%\bibliography{all}

\input{main.bbl}
\end{document}

%% file: chapters/macros.tex
\def\tref{t_\text{ref}} %
\def\pol{p} %
\def\prtl#1#2{\frac{\partial#1}{\partial#2}}
\def\tmid{t_\text{mid}} %

\def\cE{\mathcal{E}} %
\def\numEvents{N_e} %
\def\Warp{\mathbf{W}}

\def\bx{\mathbf{x}}
\def\bparams{\boldsymbol{\theta}}

\def\pol{p}
\def\velflow{\mathbf{v}}
\def\bmu{\boldsymbol{\mu}}
\def\mId{\mathtt{Id}}
\def\IWE{I}

\def\cN{\mathcal{N}} %

\def\bzero{\mathbf{0}}

\def\variance{\operatorname{Var}}

\newcommand{\unum}[2]{\multicolumn{1}{c}{\underline{\tablenum[table-format={#1}]{#2}}}}  % {arg1} ... precision, {arg2} ... value
\newcommand{\bnum}[1]{\bfseries #1}

\definecolor{light-gray}{gray}{0.6}
\newcommand\gframe[1]{{\color{light-gray}\frame{#1}}}

\def\fwd#1#2{D_{#1}^{+} #2} %
\def\bwd#1#2{D_{#1}^{-} #2}

\def\uxx{U_x}
\def\uyy{U_y}
\DeclareMathOperator{\sign}{sgn}

%% file: chapters/00_abstract.tex
\iffalse{
\begin{abstract}
Event cameras are novel sensors that naturally respond to the scene dynamics and offer advantages 
to estimate motion.
Seeking the accuracy gains that image-based deep-learning methods have achieved, 
optical flow estimation methods for event cameras have rushed to combine those image-based methods 
with ideas from top performing event-based frameworks.
However, it requires several adaptations (data conversion, loss design, etc.) 
as events and images have very different properties,
and a supervisory signal that matches the capabilities of event data.
We develop a principled method to extend the Contrast Maximization framework to estimate optical flow from events alone.
We do so by investigating key elements: 
how to design the objective function to prevent overfitting, 
how to warp events to deal better with occlusions, %
and how to improve convergence without discretizing events.
With these key elements, our method ranks first on the MVSEC indoor benchmark,
and is competitive on the MVSEC outdoor and the DSEC benchmarks.
Moreover, our method allows us to expose the issues of the ground truth flow in current datasets and benchmarks,
and produces remarkable results when it is transferred to unsupervised learning architectures.
\end{abstract}
}\fi

\begin{abstract}
Event cameras respond to scene dynamics and offer advantages to estimate motion. Following recent image-based deep-learning achievements, optical flow estimation methods for event cameras have rushed to combine those image-based methods with event data. However, it requires several adaptations (data conversion, loss function, etc.) as they have very different properties. We develop a principled method to extend the Contrast Maximization framework to estimate optical flow from events alone. We investigate key elements: how to design the objective function to prevent overfitting, how to warp events to deal better with occlusions, and how to improve convergence with multi-scale raw events. With these key elements, our method ranks first among unsupervised methods on the MVSEC benchmark, and is competitive on the DSEC benchmark. Moreover, our method allows us to expose the issues of the ground truth flow in those benchmarks, and produces remarkable results when it is transferred to unsupervised learning settings.
Our code is available at \textcolor{magenta}{\url{https://github.com/tub-rip/event_based_optical_flow}}
\end{abstract}

%% file: chapters/01_intro.tex
\section{Introduction}
\label{sec:intro}

\input{floats/fig_eye_catcher}

Event cameras are novel bio-inspired vision sensors that naturally respond to motion of edges in image space
with high dynamic range (HDR) and minimal blur at high temporal resolution (on the order of $\si{\micro\second}$) \cite{Posch14ieee,Benosman14tnnls}.
These advantages provide a rich signal for accurate motion estimation in difficult real-world scenarios for frame-based cameras.
However such a signal is, by nature, asynchronous and sparse, which is not compatible with traditional computer vision algorithms. 
This poses the challenge of rethinking visual processing \cite{Gallego20pami,Lagorce17pami}:
motion patterns (i.e., \emph{optical flow}) are no longer obtained by analyzing the intensities of images captured at regular intervals, 
but by analyzing the stream of events (per-pixel brightness changes) produced by the event camera.

Multiple methods have been proposed for event-based optical flow estimation.
They can be broadly categorized in two: 
($i$) model-based methods, which investigate the principles and characteristics of event data that enable optical flow estimation,
and ($ii$) learning-based methods, which exploit correlations in the data and/or apply the above-mentioned principles to compute optical flow. 
One of the challenges of event-based optical flow is the lack of ground truth flow in real-world datasets (at $\si{\micro\second}$ resolution and HDR) \cite{Gallego20pami}, 
which makes it difficult to evaluate and compare the methods properly, and to train supervised learning-based ones.
Ground truth (GT) in de facto standard datasets \cite{Zhu18ral,Gehrig21ral} is given by the \emph{motion field} \cite{Trucco98book} using additional depth sensors and camera information. 
However, such data is limited by the field-of-view (FOV) and resolution (spatial and temporal) of the depth sensor, which do not match those of event cameras.
Hence, it is paramount to develop interpretable optical flow methods that exploit the characteristics of event data, 
and that do not need expensive-to-collect and error-prone ground truth. %

Among prior work, Contrast Maximization (CM) \cite{Gallego18cvpr,Gallego19cvpr} is a powerful framework 
that allows us to tackle multiple motion estimation problems (rotational motion \cite{Gallego17ral,Kim21ral,Gu21iccv}, homographic motion \cite{Gallego18cvpr,Nunes21pami,Peng21pami},
feature flow estimation \cite{Zhu17icra,Zhu17cvpr,Seok20wacv,Stoffregen19cvpr}, 
motion segmentation \cite{Mitrokhin18iros,Stoffregen19iccv,Zhou21tnnls,Parameshwara21icra}, and also reconstruction \cite{Gallego18cvpr,Rebecq18ijcv,Zhang21arxiv}). 
It maximizes an objective function (e.g., contrast) that measures the alignment of events caused by the same scene edge.
The intuitive interpretation is to estimate the motion by recovering the sharp (motion-compensated) image of edge patterns that caused the events.
Preliminary work on applying CM to estimate optical flow has reported a problem of overfitting to the data, 
producing undesired flows that warp events to few pixels or lines \cite{Zhu19cvpr} (i.e., event collapse \cite{Shiba22sensors}). %
This issue has been tackled by changing the objective function, from contrast to the energy of an average timestamp image \cite{Zhu19cvpr,Paredes21cvpr,Paredes21neurips},  
but this loss is not straightforward to interpret and makes training difficult to converge \cite{Gallego19cvpr}.

Given the state-of-the-art performance of CM in low-DOF motion problems 
and its issues in more complex motions (dense flow), 
we think prior work may have rushed to use CM in unsupervised learning of complex motions.
There is a gap in understanding how CM can be sensibly extended to estimate dense optical flow accurately.
In this paper we fill this gap and learn a few ``secrets'' that are also applicable to overcome the issues of previous approaches.

We propose to extend CM for dense optical flow estimation via a tile-based approach covering the image plane.
We present several distinctive contributions:
\begin{enumerate}
    \item A \emph{multi-reference} focus loss function to improve accuracy and discourage overfitting (Sec.~\ref{sec:method:multiref}).
    \item A principled \emph{time-aware flow} to better handle occlusions, formulating event-based optical flow as a transport problem via differential equations (Sec.~\ref{sec:method:timeaware}).
    \item A \emph{multi-scale} approach on the raw events to improve convergence to the solution and avoid getting trapped in local optima (Sec.~\ref{sec:method:multiscale}).
\end{enumerate}

The results of our experimental evaluation are surprising: 
the above design choices are key to our simple, model-based tile-based method (Fig.~\ref{fig:eye_catcher}) achieving the best accuracy among all state-of-the-art methods, including supervised-learning ones, on the de facto benchmark of MVSEC indoor sequences \cite{Zhu18rss}.
Since our method is interpretable and produces better event alignment than the ground truth flow, both qualitatively and quantitatively, the experiments also expose the limitations of the current ``ground truth''.
Finally, experiments demonstrate that the above key choices are transferable to unsupervised learning methods,
thus guiding future design and understanding of more proficient Artificial Neural Networks (ANNs) for event-based optical flow estimation.

Because of the above, we believe that the proposed design choices deserve to be called ``secrets'' \cite{Sun13ijcv}.
To the best of our knowledge, they are novel in the context of event-based optical flow estimation, 
e.g., no prior work considers constant flow along its characteristic lines, 
designs the multi-reference focus loss to tackle overfitting, 
or has explicitly defined multi-scale (i.e., multi-resolution) contrast maximization on the raw events. 

%% file: floats/fig_eye_catcher.tex
\global\long\def\figWidth{0.24\linewidth}
\begin{figure*}[t]
	\centering
    {\scriptsize
    \setlength{\tabcolsep}{2pt}
	\begin{tabular}{
	>{\centering\arraybackslash}m{\figWidth} 
	>{\centering\arraybackslash}m{\figWidth} 
	>{\centering\arraybackslash}m{\figWidth} 
	>{\centering\arraybackslash}m{\figWidth}}
	
		\gframe{\includegraphics[width=\linewidth]{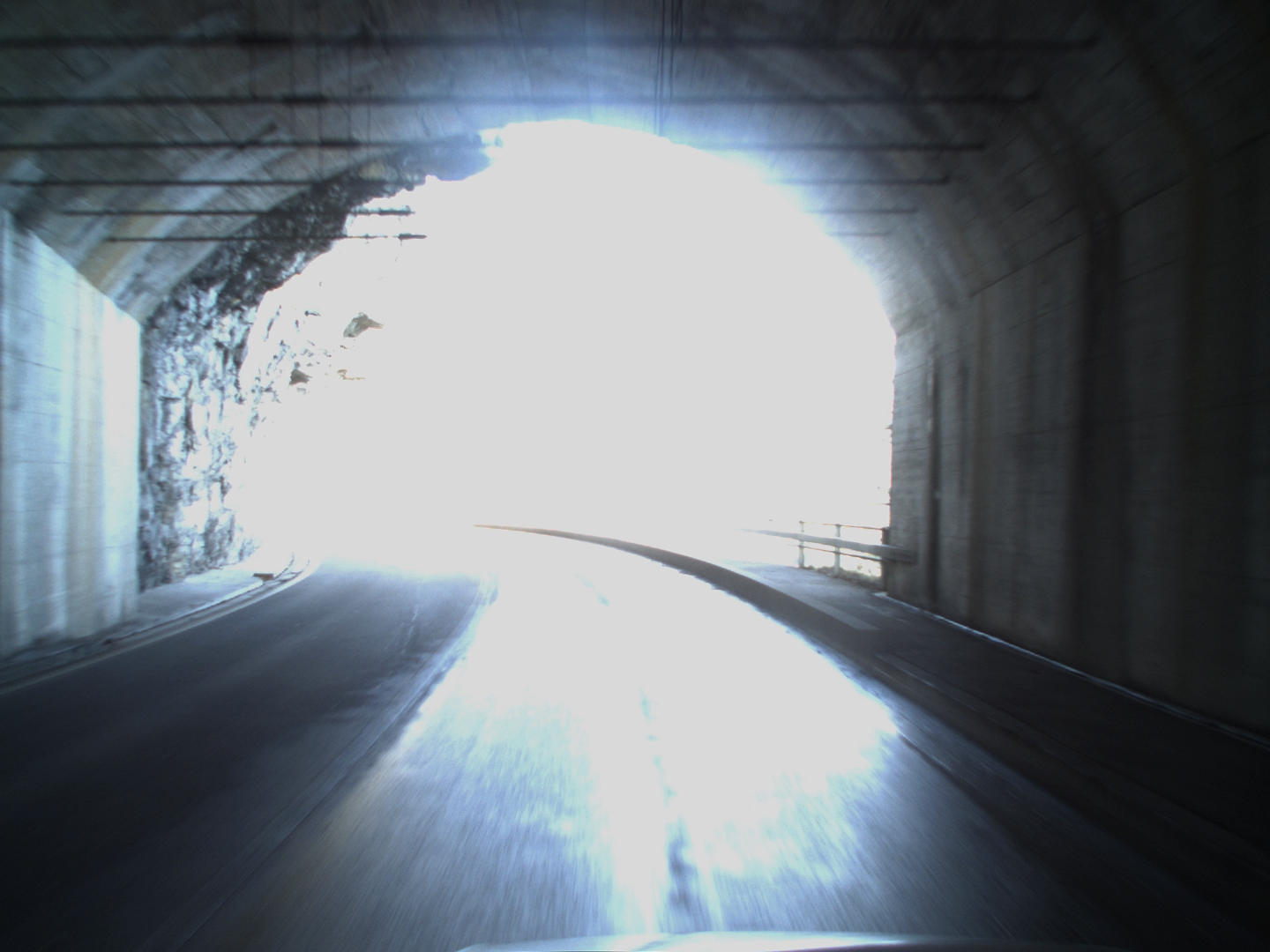}}
		&\gframe{\includegraphics[width=\linewidth]{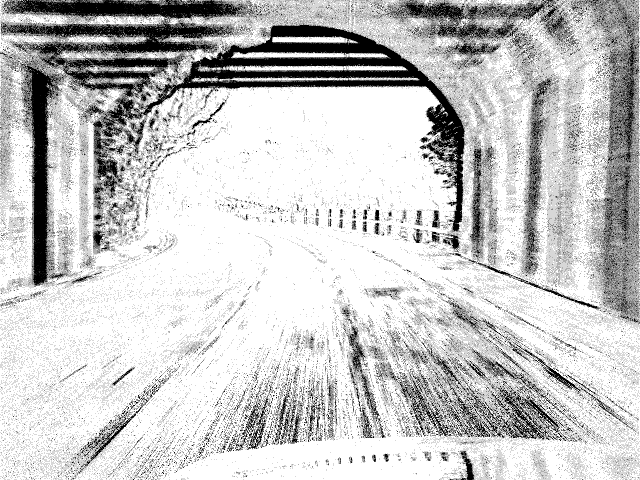}}
		&\gframe{\includegraphics[width=\linewidth]{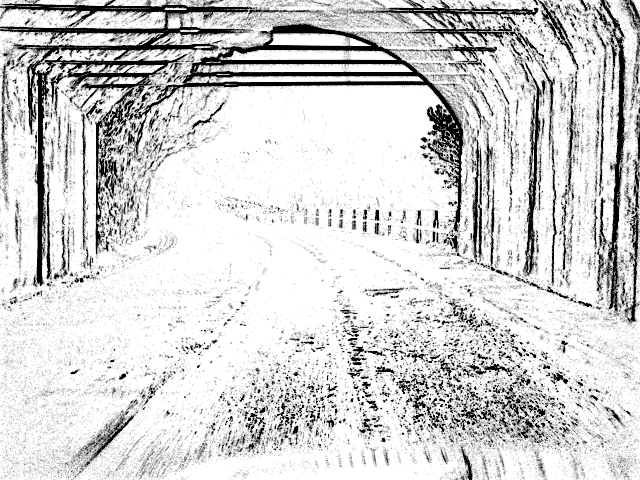}}
		&\gframe{\includegraphics[width=\linewidth]{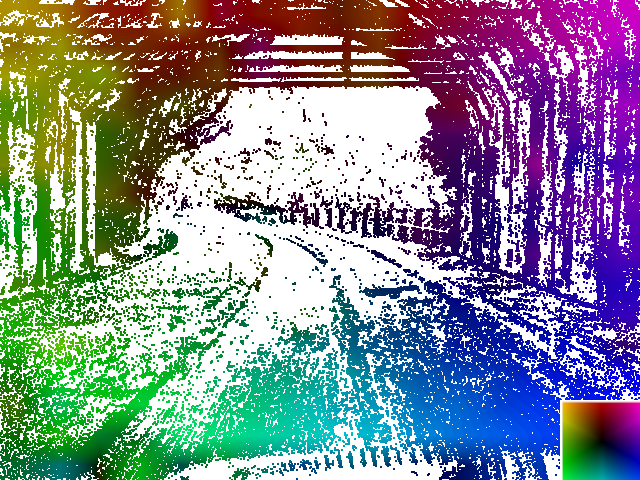}}\\

		\gframe{\includegraphics[width=\linewidth]{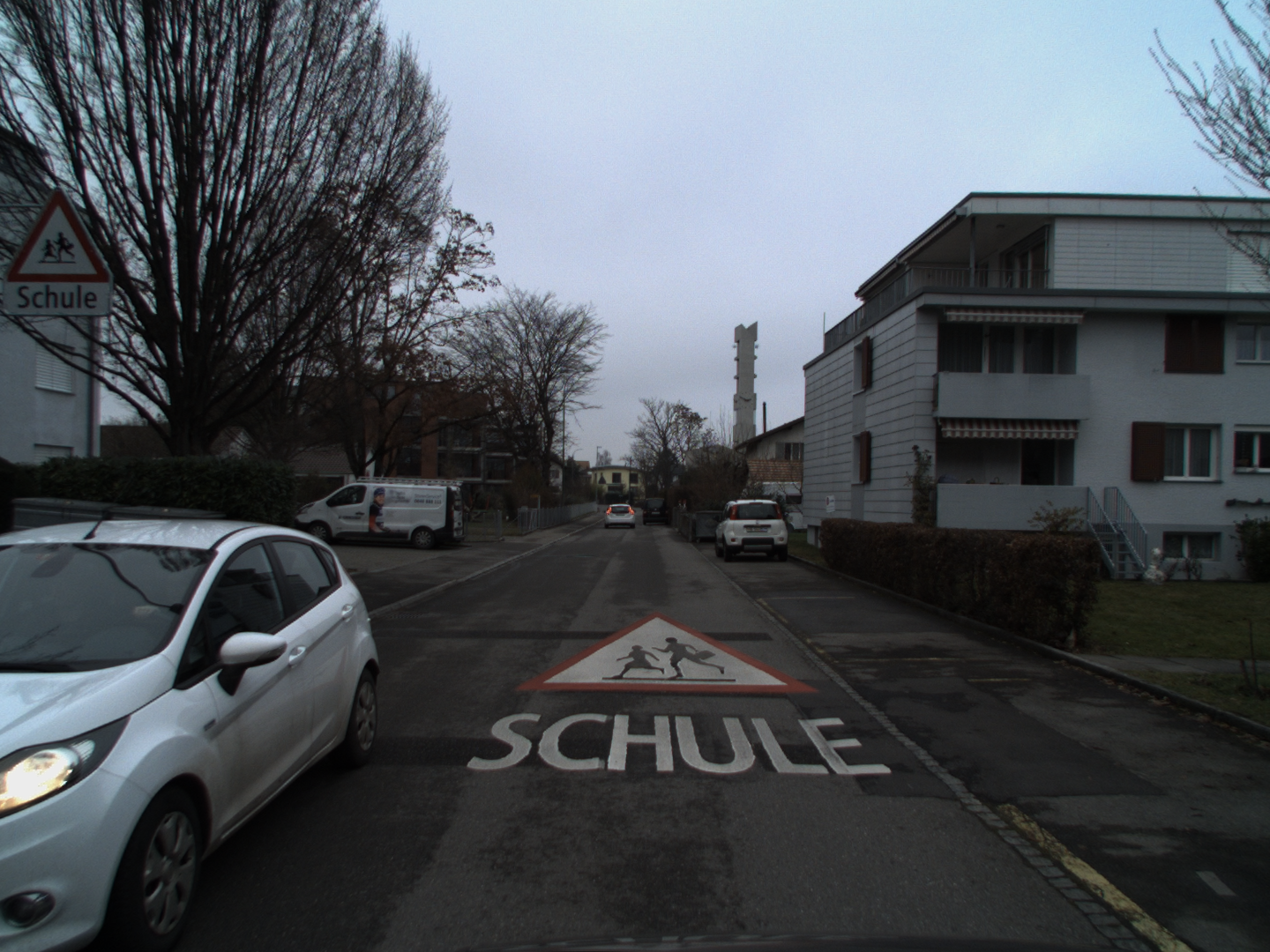}}
		&\gframe{\includegraphics[width=\linewidth]{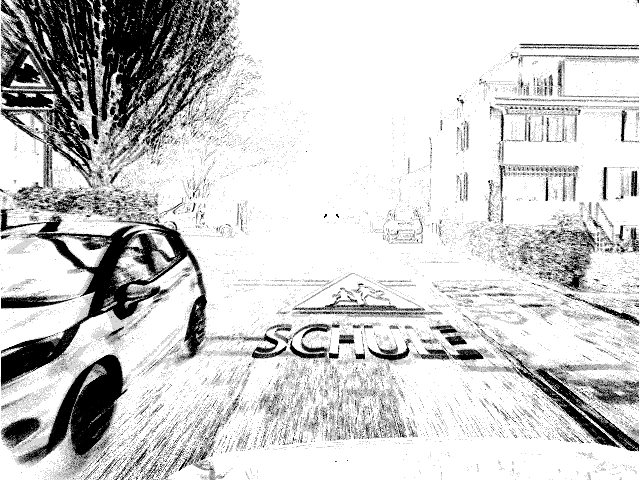}}
		&\gframe{\includegraphics[width=\linewidth]{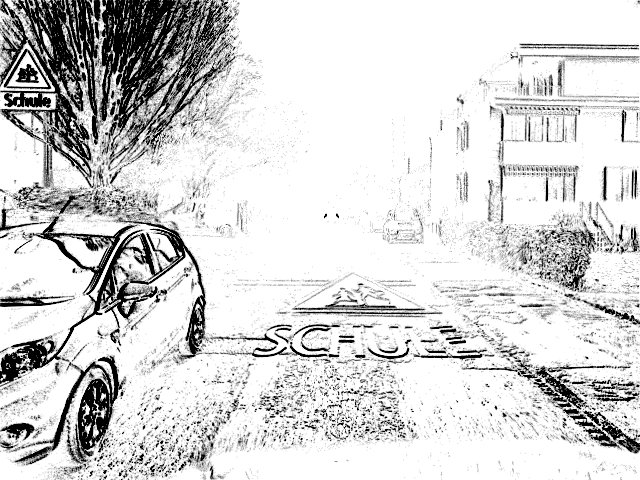}}
		&\gframe{\includegraphics[width=\linewidth]{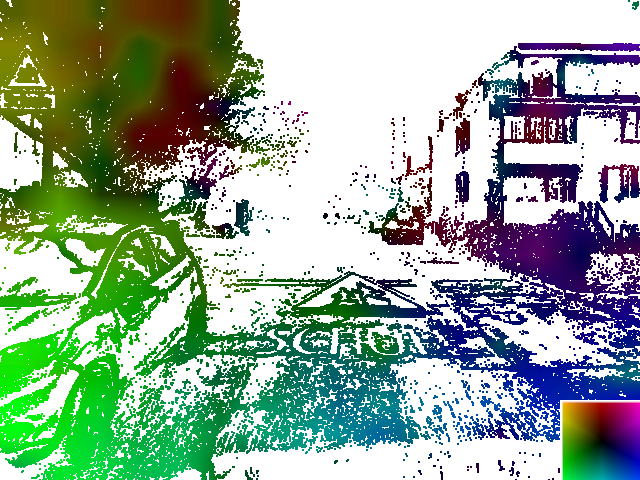}}\\
		
		(a) Frame (reference)
		& (b) Events (zero flow)
		& (c)\! Warped events, IWE
		& (d) Estimated flow\\
	\end{tabular}
	}
	\caption{Two test sequences (interlaken\_00\_b, thun\_01\_a) 
	from the DSEC dataset \cite{Gehrig21ral}. 
	Our optical flow estimation method produces sharp images of warped events (IWE) despite the scene complexity, the large pixel displacement and the high dynamic range. 
	The examples utilize 500k events on an event camera with $640 \times 480$ pixels.
	}
\label{fig:eye_catcher}
\end{figure*}

%% file: chapters/02_related.tex
\section{Prior Work on Event-based Optical Flow Estimation}
\label{sec:related}

Given the identified advantages of event cameras to estimate optical flow, 
extensive research on this topic has been carried out. %
Prior work has proposed adaptations of frame-based approaches (block matching \cite{Liu18bmvc}, Lucas-Kanade \cite{Benosman12nn}),
filter-banks \cite{Orchard13biocas,Brosch15fns},
spatio-temporal plane-fitting \cite{Benosman14tnnls,Akolkar22pami}, 
time surface matching \cite{Nagata21sensors},
variational optimization on voxelized events \cite{Bardow16cvpr},
and feature-based contrast maximization \cite{Zhu17icra,Stoffregen17acra,Gallego18cvpr}.
For a detailed survey, we refer to~\cite{Gallego20pami}.

Current state-of-the-art approaches are ANNs \cite{Zhu18rss,Zhu19cvpr,Gehrig21threedv,Ziluo21arxiv,Lee20eccv,Paredes21neurips}, 
largely inspired by frame-based optical flow architectures \cite{Ronneberger15icmicci,Teed20eccv}.
Non-spiking--based approaches need to additionally adapt the input signal, converting the events into a tensor representation (event frames, time surfaces, voxel grids, etc. \cite{Gehrig19iccv,Cannici20eccv}).
These learning-based methods can be classified into supervised, semi-supervised or unsupervised (to be used in Tab.~\ref{tab:main}).
In terms of architectures, the three most common ones are U-Net \cite{Zhu18rss}, FireNet \cite{Zhu19cvpr} and RAFT \cite{Gehrig21threedv}.

Supervised methods train ANNs in simulation and/or real-data \cite{Gehrig21threedv,Stoffregen20eccv,Gehrig19iccv}.
This requires accurate ground truth flow that matches the space-time resolution of event cameras. 
While this is no problem in simulation, it incurs in a performance gap when trained models are used to predict flow on real data \cite{Stoffregen20eccv}. 
Besides, real-world datasets have issues in providing accurate ground truth flow.

Semi-supervised methods use the grayscale images from a colocated camera (e.g., DAVIS \cite{Taverni18tcsii}) as a supervisory signal: images are warped using the flow predicted by the ANN and their photometric consistency is used as loss function \cite{Zhu18rss,Lee20eccv,Ziluo21arxiv}. 
While such supervisory signal is easier to obtain than real-world ground truth flow, it may suffer from the limitations of frame-based cameras (e.g., low dynamic range and motion blur), consequently affecting the trained ANNs.
These approaches were pioneered by EV-FlowNet \cite{Zhu18rss}.

Unsupervised methods rely solely on event data. 
Their loss function consists of an event alignment error using the flow predicted by the ANN \cite{Zhu19cvpr,Paredes21cvpr,Paredes21neurips,Ye19arxiv}. 
Zhu et al. \cite{Zhu19cvpr} extended EV-FlowNet \cite{Zhu18rss} to the unsupervised setting using a motion-compensation loss function in terms of average timestamp images. 
This approach has been used and improved in \cite{Paredes21cvpr,Paredes21neurips}.
Paredes-Vall\'es et al. \cite{Paredes21cvpr} also proposed FireFlowNet, a lightweight ANN producing competitive results.

Similar to the works in the unsupervised category,
our method produces dense optical flow and does not need ground truth or additional supervisory signals.
In contrast to prior work, we adopt a more classical modeling perspective to gain insight into the problem 
and discover principled solutions that can subsequently be applied to the learning-based setting. 
Stemming from an accurate and spatially-dependent contrast loss (the gradient magnitude \cite{Gallego19cvpr}), 
we model the problem using a tile of patches and propose solutions to several problems: 
overfitting, occlusions and convergence. 
To the best of our knowledge, ($i$) no prior work has proposed to estimate dense optical flow from a CM model-based perspective,
and ($ii$) no prior unsupervised learning approach based on motion compensation has succeeded in estimating optical flow without the average timestamp image loss.

%% file: chapters/03_method.tex
\section{Method}
\label{sec:method}

\subsection{Event Cameras and Contrast Maximization}
\label{sec:method:cmaxreview}
Event cameras have independent pixels that operate continuously and generate ``events'' $e_k \doteq (\bx_k,t_k,\pol_k)$ whenever the logarithmic brightness at the pixel increases or decreases by a predefined amount, called contrast sensitivity. 
Each event $e_k$ contains the pixel-time coordinates ($\bx_k, t_k$) of the brightness change and its polarity $\pol_k=\{+1,-1\}$. %
Events occur asynchronously %
and sparsely on the pixel lattice, with a variable rate that depends on the scene dynamics.

The CM framework \cite{Gallego18cvpr} assumes events $\cE \doteq \{e_k\}_{k=1}^{\numEvents}$
are generated by moving edges, and transforms them geometrically
according to a motion model $\Warp$, producing a set of warped events $\cE'_{\tref} \doteq \{e'_k\}_{k=1}^{\numEvents}$ at a reference time $\tref$:
\begin{equation}
e_k \doteq (\bx_k,t_k,\pol_k) \;\,\mapsto\;\, 
e'_k \doteq (\bx'_k,\tref,\pol_k).
\end{equation}
The warp $\bx'_k = \Warp(\bx_k,t_k; \bparams)$ transports each event from $t_k$ to $\tref$ along the motion curve that passes through it. 
The vector $\bparams$ parametrizes the motion curves.
Transformed events are aggregated on an image of warped events (IWE):
\begin{equation}
\label{eq:IWE}
\textstyle
\IWE(\bx; \cE'_{\tref}, \bparams) \doteq \sum_{k=1}^{\numEvents} \delta (\bx - \bx'_k),
\end{equation}
where each pixel $\bx$ sums the number of warped events $\bx'_k$ that fall within it.
The Dirac delta $\delta$ is approximated by a Gaussian, 
$\delta(\bx-\bmu)\approx\cN(\bx;\bmu,\epsilon^2\mId)$ with $\epsilon=1$ pixel.
Next, an objective function $f(\bparams)$ is built from the transformed events, 
such as the contrast of the IWE~\eqref{eq:IWE}, given by the variance
\begin{equation}
\label{eq:IWEVariance}
\textstyle
\variance\bigl(\IWE(\bx;\bparams)\bigr) 
\doteq \frac{1}{|\Omega|} \int_{\Omega} (\IWE(\bx;\bparams)-\mu_{\IWE})^2 d\bx,
\end{equation}
with mean $\mu_{\IWE} \doteq \frac{1}{|\Omega|} \int_{\Omega} \IWE(\bx;\bparams) d\bx$.
The objective function measures the goodness of fit between the events and the candidate motion curves (warp).
Finally, an optimization algorithm %
iterates the above steps 
until convergence.
The goal is to find the motion parameters %
that maximize the alignment of events caused by the same scene edge.
Event alignment is measured by the strength of the edges of the IWE, 
which is directly related to image contrast \cite{Gonzalez09book}.

\textbf{Dense optical flow}. In the task of interest, the warp used is \cite{Zhu19cvpr,Paredes21cvpr,Paredes21neurips}:
\begin{equation}
\label{eq:warp:oflow}
\bx'_k = \bx_k + (t_k-\tref) \, \velflow(\bx_k),
\end{equation}
where $\bparams = \{ \velflow(\bx) \}_{\bx\in\Omega}$ is a flow field on the image plane at a set time, e.g., $\tref$.

\subsection{Multi-reference Focus Objective Function}
\label{sec:method:multiref}

Zhu et al. \cite{Zhu19cvpr} report that the contrast objective (variance) overfits to the events.
This is in part because the warp \eqref{eq:warp:oflow} can describe very complex flow fields, 
which can push the events to accumulate in few pixels \cite{Shiba22sensors}.  
To mitigate overfitting, we reduce the complexity of the flow field by dividing the image plane into a tile of non-overlapping patches, 
defining a flow vector at the center of each patch and interpolating the flow on all other pixels
(we show the tiles in Sec.~\ref{sec:method:multiscale}).

However, this is not enough.
Additionally, we discover that warps that produce sharp IWEs \emph{at any} reference time $\tref$ have a regularizing effect on the flow field, discouraging overfitting.
This is illustrated in Fig.~\ref{fig:fig_method_flow}.
In practice we compute the \emph{multi-reference} focus loss using 3 reference times: 
$t_1$ (min), $\tmid\doteq (t_1+t_{\numEvents})/2$ (midpoint) and $t_{\numEvents}$ (max).
The flow field is defined only at one reference time.
\input{floats/fig_method_flow}

Furthermore, we measure event alignment using the magnitude of the IWE gradient because: 
($i$) it has top accuracy performance among the objectives in \cite{Gallego19cvpr},
($ii$) it is sensitive to the arrangement (i.e., permutation) of the IWE pixel values,
whereas the variance of the IWE \eqref{eq:IWEVariance} is not,
($iii$) it converges more easily than other objectives we tested,
($iv$) it differs from the Flow Warp Loss (FWL) \cite{Stoffregen20eccv}, which is defined using the variance \eqref{eq:IWEVariance} and will be used for evaluation.

Finally, letting the (squared) gradient magnitude of the IWE be  
\begin{equation}
\label{eq:gradmag}
\textstyle
    G(\bparams; \tref) \doteq \frac{1}{|\Omega|} \int_{\Omega} \| \nabla \IWE(\bx; \tref)\|^2\,d\bx ,
\end{equation}
the proposed multi-reference focus objective function becomes the average of the $G$ functions of the IWEs at multiple reference times:
\begin{equation}
\label{eq:theloss}
    f(\bparams) \doteq \bigl(G(\bparams; t_1) + 2G(\bparams; \tmid) + G(\bparams; t_{\numEvents})\bigr) \,/\, 4 G(\bzero; -),
\end{equation}
normalized by the value of the $G$ function with zero flow (identity warp).
The normalization in \eqref{eq:theloss} provides the same interpretation as the FWL: 
$f<1$ implies the flow is worse than the zero flow baseline, 
whereas $f>1$ means that the flow produces sharper IWEs than the baseline.

\emph{Remark}: Warping to two reference times (min and max) was proposed in \cite{Zhu19cvpr}, 
but with important differences: 
($i$) it was done for the average timestamp loss, 
hence it did not consider the effect on contrast or focus functions \cite{Gallego19cvpr}, 
and ($ii$) it had a completely different motivation: 
to lessen a back-propagation scaling problem, 
so that the gradients of the loss would not favor events far from $\tref$.

\subsection{Time-aware Flow}
\label{sec:method:timeaware}

State-of-the-art event-based optical flow approaches are based on frame-based ones, 
and so they use the warp \eqref{eq:warp:oflow}, which defines the flow $\velflow(\bx)$ as a function of $\bx$
(i.e., a pixel displacement between two given frames).
However, this does not take into account the space-time nature of events, which is the basis of the CM approach, because not all events at a pixel $\bx_0$ are triggered at the same timestamp $t_k$.
They do not need to be warped with the same velocity $\velflow(\bx_0)$.
Figure \ref{fig:fig_method_timeaware} illustrates this with an occlusion example taken from the slider\_depth sequence \cite{Mueggler17ijrr}.
Instead of $\velflow(\bx)$, the \emph{event-based flow} should be a function of space-time, $\velflow(\bx,t)$, i.e, \emph{time-aware}, 
and each event $e_k$ should be warped according to the flow defined at ($\bx_k,t_k$).
Let us propose a more principled warp than \eqref{eq:warp:oflow}.

\input{floats/fig_method_timeaware}
To define a space-time flow $\velflow(\bx,t)$ that is compatible with the propagation of events along motion curves, 
we are inspired by the method of characteristics \cite{Evans10book}. 
Just like the brightness constancy assumption states that brightness is constant along the true motion curves in image space, 
we assume the flow is constant along its streamlines: $\velflow(\bx(t),t) = \text{const}$ (Fig.~\ref{fig:fig_method_timeaware}).
Differentiating in time and applying the chain rule gives a system of partial differential equations (PDEs):
\begin{equation}
    \label{eq:flow:pde}
    \prtl{\velflow}{\bx} \frac{d\bx}{dt} + \prtl{\velflow}{t} = \bzero,
\end{equation}
where, as usual, $\velflow = d\bx/dt$ is the flow.
The boundary condition is given by the flow at say $t=0$: $\velflow(\bx,0) = \velflow^0(\bx)$.
This system of PDEs essentially states how to propagate (i.e., \emph{transport}) a given flow $\velflow^0(\bx)$, from the boundary $t=0$ to the rest of space $\bx$ and time $t$.
The PDEs have advection terms and others that resemble those of the inviscid Burgers' equation \cite{Evans10book} since the flow is transporting itself.
We parametrize the flow at $t=\tmid$ (boundary condition), and then propagate it to the volume that encloses the current set of events $\cE$.
We develop two explicit methods to solve the PDEs, one with upwind differences and one with a conservative scheme adapted to Burgers' terms \cite{Sethian99book}. %
Each event $e_k$ is then warped according to a flow $\hat{\velflow}$ given by the solution of the PDEs at ($\bx_k,t_k$): %
\begin{equation}
\label{eq:warp:oflow:PDE}
\bx'_k = \bx_k + (t_k-\tref) \, \hat{\velflow}(\bx_k,t_k).
\end{equation}

\subsection{Multi-scale Approach}
\label{sec:method:multiscale}

Inspired by classical estimation methods, we combine our tile-based approach with a multi-scale strategy. 
The goal is to improve the convergence of the optimizer in terms of speed and robustness (i.e., avoiding local optima).

Some learning-based works \cite{Zhu18rss,Zhu19cvpr,Paredes21cvpr} also have a multi-scale component, inherited from the use of a U-Net architecture.
However, they work on discretized event representations (voxel grid, etc.) to be compatible with CNNs.
In contrast, our tile-based approach works directly on raw events, without discarding or quantizing the temporal information in the event stream. 
While some prior work outside the context of optical flow has considered multi-resolution on raw events
\cite{Li19neucom},  %
there is no agreement on the best way to perform multi-resolution due to the sparse and asynchronous nature of events.

\input{floats/fig_method_multi_scale}

Our multi-scale CM approach is illustrated in Fig.~\ref{fig:fig_method_multi_scale}.
For an event set $\cE_i$, we apply the tile-based CM in a coarse-to-fine manner (e.g., $N_{\ell} = 5$ scales).
There are $2^{l - 1} \times 2^{l - 1}$ tiles at the $l$-th scale.
We use bilinear interpolation to upscale between any two scales.
If there is a subsequent set $\cE_{i+1}$, the flow estimated from $\cE_{i}$ is used to initialize the flow for $\cE_{i+1}$.
This is done by downsampling the finest flow to coarser scales. The coarsest scale initializes the flow for $\cE_{i+1}$. For finer scales, initialization is computed as the average of the upsampled flow from the coarser scale of $\cE_{i+1}$ and the same-scale flow from $\cE_{i}$.

\textbf{Composite Objective}.
\label{sec:method:composite}
To encourage additional smoothness of the flow, even in regions with few events, %
we include a flow regularizer $\mathcal{R}(\bparams)$. 
The flow is obtained as the solution to the problem with the composite objective:
\begin{equation}
\label{eq:composite:function}
\bparams^{\ast} = \arg\min_{\bparams} \bigl(1 / f(\bparams) + \lambda \mathcal{R}(\bparams)\bigr),
\end{equation}
where, $\lambda>0$ is the regularizer weight, and we use the total variation (TV) \cite{Rudin92physica} as regularizer.
We choose $1/f$ instead of simply $-f$ because it is convenient for ANN training, as we will apply in Sec \ref{sec:experim:applytodnn}.

%% file: floats/fig_method_flow.tex
\begin{figure*}[t]
	\centering
	\includegraphics[trim={5cm 5cm 1cm 7.1cm},clip, width=0.8\linewidth]{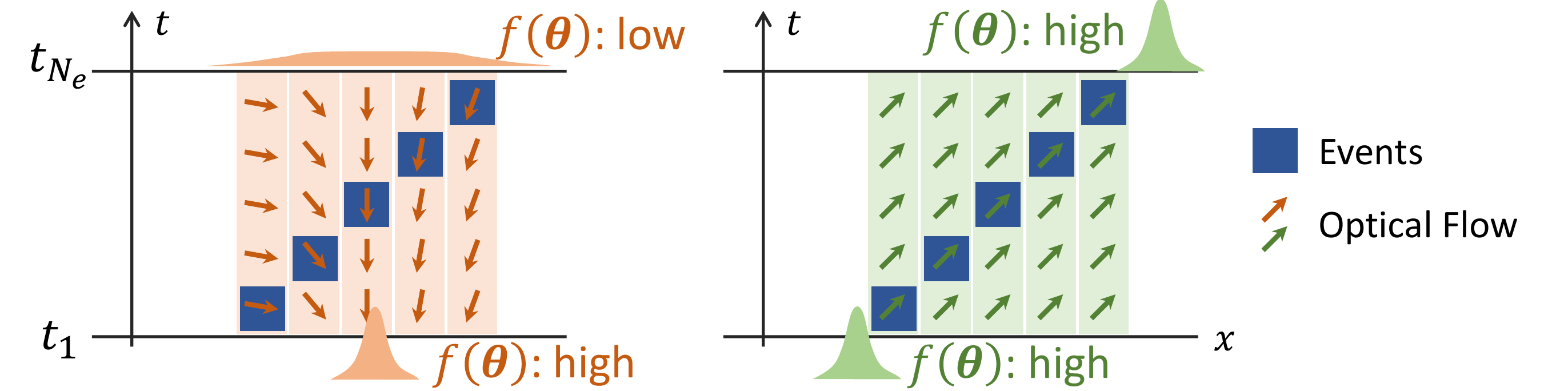}
	\caption{\emph{Multi-reference focus loss}.
	Assume an edge moves from left to right.
	Flow estimation with single reference time ($t_1$) can overfit to the data, warping all events into a single pixel, which results in a maximum contrast (at $t_1$).
	However, the same flow would produce low contrast (i.e., a blurry image) if events were warped to time $t_{\numEvents}$.
	Instead, we favor flow fields that produce high contrast (i.e., sharp images) at any reference time (here, $\tref = t_1$ and $\tref = t_{\numEvents}$).
	See results in Fig. \ref{fig:suppl_abl_multiref}.
	}
\label{fig:fig_method_flow}
\end{figure*}

%% file: floats/fig_method_timeaware.tex
\global\long\def\figWidth{\linewidth}
\begin{figure*}[t]
	\centering
	\includegraphics[trim={0 4.6cm 0 7.8cm},clip, width=\linewidth]{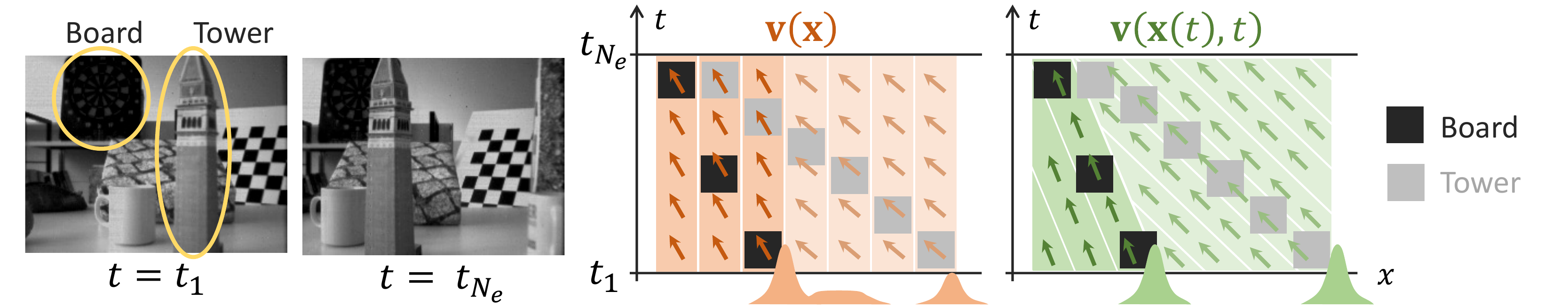}
	\caption{\emph{Time-aware Flow.}
	Traditional flow \eqref{eq:warp:oflow}, inherited from frame-based approaches, assumes per-pixel constant flow $\velflow(\bx) = \text{const}$, which cannot handle occlusions properly. 
	The proposed space-time flow assumes constancy along streamlines, $\velflow(\bx(t),t) = \text{const}$, which allows us to handle occlusions more accurately. (See results in Fig. \ref{fig:effect_of_time_aware})}
\label{fig:fig_method_timeaware}
\end{figure*}

%% file: floats/fig_method_multi_scale.tex
\global\long\def\figWidth{1.0\linewidth}
\begin{figure*}[t]
	\centering
	{\includegraphics[trim={0 3.8cm 0 5cm},clip, width=0.9\linewidth]{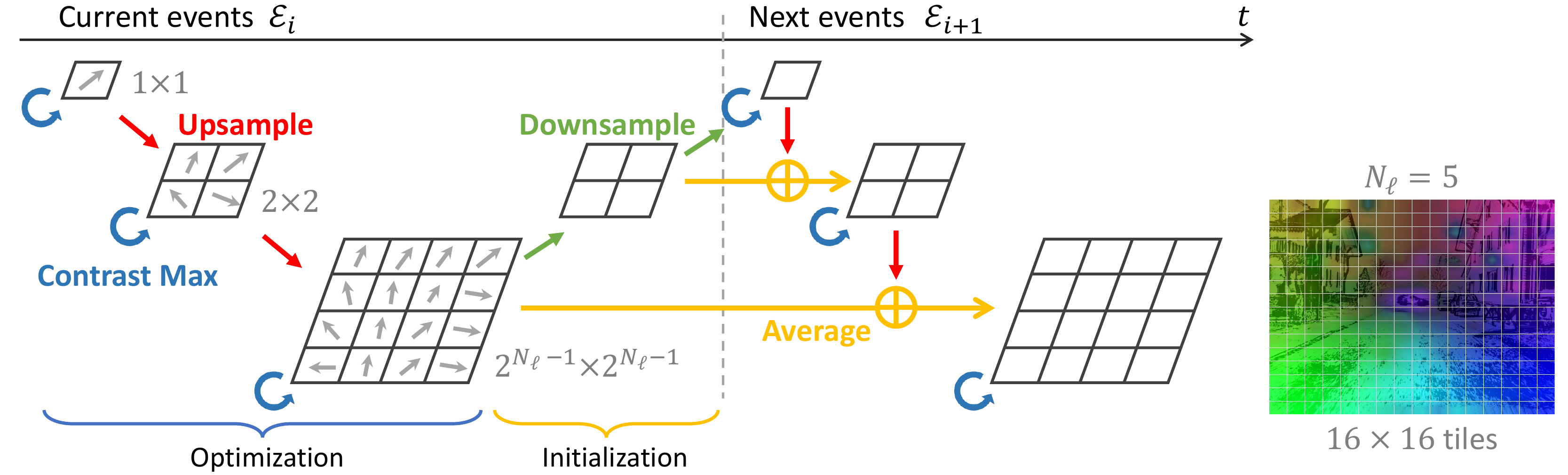}}
	\caption{\emph{Multi-scale Approach} using tiles (rectangles) and raw events.}
\label{fig:fig_method_multi_scale}
\end{figure*}

%% file: chapters/04_experiments.tex
\section{Experiments}
\label{sec:experim}

\subsection{Datasets, Metrics and Hyper-parameters}
\label{sec:experim:datasets}

We evaluate our method on sequences from the MVSEC dataset \cite{Zhu18ral,Zhu18rss}, which is the de facto dataset used by prior works to benchmark optical flow. 
It provides events, grayscale frames, IMU data, camera poses, and scene depth from a LiDAR \cite{Zhu18ral}. 
The dataset was extended in \cite{Zhu18rss} to provide ground truth optical flow, computed as the motion field \cite{Trucco98book} given the camera velocity and the depth of the scene.
The event camera has $346 \times 260$ pixel resolution \cite{Taverni18tcsii}. 
In total, we evaluate on 63.5 million events spanning 265 seconds.

We also evaluate on a recent dataset that provides ground truth flow: DSEC \cite{Gehrig21threedv}.
It consists of sequences recorded with Prophesee Gen3 event cameras, of higher resolution ($640 \times 480$ pixels), mounted on a car.
Optical flow is also computed as the motion field, with the scene depth from a LiDAR.
In total, we evaluate on 3 billion events spanning the 208 seconds of the test sequences.

The metrics used to assess optical flow accuracy are 
the average endpoint error (AEE) and the percentage of pixels with AEE greater than 3 pixels %
(denoted by ``\% Out''), 
both are measured over pixels with valid ground-truth and at least one event in the evaluation intervals.
We also use the FWL metric (the IWE variance relative to that of the identity warp) to assess event alignment  \cite{Stoffregen20eccv}.

In all experiments our method uses $N_{\ell}=5$ resolution scales, %
$\lambda = 0.0025$ in \eqref{eq:composite:function}, 
and the Newton-CG optimization algorithm with a maximum of 20 iterations/scale. 
The flow at $\tmid$ is transported to each side via the upwind or Burgers' PDE solver 
(using 5 bins for MVSEC, 40 for DSEC), 
and used for event warping \eqref{eq:warp:oflow:PDE} (see Suppl. Mat.).
In the optimization, we use 30k events for MVSEC indoor sequences, 40k events for outdoors, and $1.5$M events for DSEC.

\subsection{Results on MVSEC}
\label{sec:experim:mvsec}
\input{floats/tab_main}
\input{floats/fig_qualitative_large}

Table \ref{tab:main} reports the results on the MVSEC benchmark.
The different methods (rows) are compared on three indoor sequences and one outdoor sequence (columns). 
This is because many learning-based methods train on the other outdoor sequence, which is therefore not used for testing. 
Following Zhu et al., outdoor\_day1 is tested only on specified 800 frames \cite{Zhu18rss}.
The top part of Tab.~\ref{tab:main} reports the flow corresponding to a time interval of $dt=1$ grayscale frame (at $\approx$ 45Hz, i.e., 22.2ms),
and the bottom part corresponds to $dt=4$ frames (89ms).

Our methods provide the best results among all methods in all indoor sequences 
and are the best among the unsupervised and model-based methods in the outdoor sequence.
The errors for $dt=4$ are about four times larger than those for $dt=1$, which is sensible given the ratio of time interval sizes.
We observe no significant differences between the three versions of the method tested (warp models, 
see also Sec. \ref{sec:experim:timeaware}), 
which can be attributed to the fact that the MVSEC dataset does not comprise large pixel displacements or occlusions.

Qualitative results are shown in Fig.~\ref{fig:qualitative_comparison}, 
where we compare our method %
against the state of the art. 
Our method provides sharper IWEs than the baselines, without overfitting, 
and the estimated flow resembles the ground truth one.

Ground truth (GT) is not available on the entire image plane (see Fig.~\ref{fig:qualitative_comparison}), 
such as in pixels not covered by the LiDAR's range, FOV, or spatial sampling.
Additionally, there may be interpolation issues in the GT, since the LiDAR works at 20 Hz and the GT flow is given at frame rate (45 Hz).
In the outdoor sequences, the GT from the LiDAR and the camera motion cannot provide correct flow for independently moving objects (IMOs). %
These issues of the GT are
noticeable in the IWEs: they are not as sharp as expected.
In contrast, the IWEs produced by our method are sharp.
Taking now into account the GT quality on the comparison Table \ref{tab:main},
it is remarkable that our method outperforms the state-of-the-art baselines on the indoor sequences, 
where GT has the best quality (with more points in the valid LiDAR range and no IMOs).

\subsection{Results on DSEC}
\label{sec:experim:dsec}

Table~\ref{tab:dsec} gives quantitative results on the DSEC Optical Flow benchmark.
Currently only the method that proposed the benchmark reports values \cite{Gehrig21threedv}.
As expected, this supervised learning method is better than ours in terms of flow accuracy 
because ($i$) it has additional training information (GT labels), 
and ($ii$) it is trained using the same type of GT signal used in the evaluation.
Nevertheless, our method provides competitive results and is better in terms of FWL, which exposes similar GT quality issues as those of MVSEC:
pixels without GT (LiDAR's FOV and IMOs).
Qualitative results are shown in Fig.~\ref{fig:qualitative_dsec}.
Our method provides sharp IWEs, even for IMOs (car) and the road close to the camera.

\input{floats/tab_dsec}

\input{floats/fig_qualitative_dsec}

We observe that the evaluation intervals (100ms) are large for optical flow standards.
In the benchmark, 80\% of the GT flow has up to 22px displacement, 
which means that 20\% of the GT flow is larger than 22px (on VGA resolution).
The apparent motion during such intervals is sufficiently large that it breaks the classical assumption of scene points flowing in linear trajectories.

\subsection{Effect of the Multi-reference Focus Loss}

The effect of the proposed multi-reference focus loss is shown in Fig.~\ref{fig:suppl_abl_multiref}.
The single-reference focus loss function can easily overfit to the only reference time, pushing all events into a small region of the image 
at $t_1$ while producing blurry IWEs at other times ($\tmid$ and $t_{\numEvents}$).
Instead, our proposed multi-reference focus loss discourages such overfitting, as the loss favors flow fields which produce sharp IWEs at \emph{any} reference time.
The difference is also noticeable in the flow: the flow from the single-reference loss is irregular, with a lot of spatial variability in terms of directions (many colors, often in opposite directions of the color wheel). 
In contrast, the flow from the multi-reference loss is considerably more regular.

\input{floats/fig_suppl_abl_multiref}

\subsection{Effect of the Time-Aware Flow}
\label{sec:experim:timeaware}

To assess the effect of the proposed time-aware warp \eqref{eq:warp:oflow:PDE}, we conducted experiments on MVSEC, DSEC and ECD~\cite{Mueggler17ijrr} datasets.
Accuracy results are already reported in Tabs. \ref{tab:main} and \ref{tab:dsec}.
We now report values of the FWL metric in Tab.~\ref{tab:fwl}.
For MVSEC, $dt=1$ is a very short time interval, with small motion and therefore few events, hence the sharpness of the IWE with or without motion compensation are about the same (FWL $\approx 1$).
Instead, $dt=4$ provides more events, and larger FWL values (1.1--1.3), which means that the contrast of the motion-compensated IWE is better than that of the zero flow baseline. %
All three methods provide sharper IWEs than ground truth. 
The advantages of the time-aware warp \eqref{eq:warp:oflow:PDE} over \eqref{eq:warp:oflow} to produce better IWEs (higher FWL) are most noticeable on sequences like slider\_depth \cite{Mueggler17ijrr} and DSEC (see Fig.~\ref{fig:effect_of_time_aware}) because of the occlusions and larger motions.
Notice that FWL differences below 0.1 are significant \cite{Stoffregen20eccv}, demonstrating the efficacy of time-awareness.

\input{floats/tab_fwl}
\input{floats/fig_effect_of_time_awareness}

\subsection{Application to Deep Neural Networks (DNN)}
\label{sec:experim:applytodnn}

The proposed secrets are not only applicable to model-based methods, but also to unsupervised-learning methods.
We train EV-FlowNet \cite{Zhu18rss} in an unsupervised manner, using \eqref{eq:composite:function} as data-fidelity term and a Charbonnier loss \cite{Charbonnier97tip} as the regularizer.
Since the time-aware flow does not have a significant influence on the MVSEC benchmark (Tab.~\ref{tab:main}), we do not port it to the learning-based setting.
We convert 40k events into the voxel-grid representation \cite{Zhu19cvpr} with 5 time bins.
The network is trained for 50 epochs with a learning rate of $0.001$ with Adam optimizer and with 0.8 learning rate decay.
To ensure generalization, we train our network on indoor sequences and test on the outdoor\_day1 sequence.

\input{floats/tab_dnn}

Table~\ref{tab:dnn} shows the quantitative comparison with unsupervised-learning methods.
Our model achieves the second best accuracy, following \cite{Zhu19cvpr}, and the best sharpness (FWL) among the existing methods.
Notice that \cite{Zhu19cvpr} was trained on the outdoor\_day2 sequence,
which is a similar driving sequence to the test one,
while the other methods were trained on drone data \cite{Delmerico19icra}.
Hence \cite{Zhu19cvpr} might be overfitting to the driving data, while ours is not,
by the choice of training data.

%% file: floats/tab_main.tex
\begin{table}[t]
\centering
\caption{Results on MVSEC dataset \cite{Zhu18rss}.
Methods are sorted according to how much data they need: 
supervised learning (SL) requires ground truth flow;
semi-supervised learning (SSL) uses grayscale images for supervision;
unsupervised learning (USL) uses only events;
and model-based (MB) needs no training data.
Bold is the best among all methods; 
underlined is second best.
Nagata et al.~\cite{Nagata21sensors} evaluate on shorter time intervals; for comparison, we scale the errors to $dt=1$.
}
\label{tab:main}
\adjustbox{max width=\textwidth}{%
\setlength{\tabcolsep}{2pt}
\begin{tabular}{ll*{8}{S[table-format=2.3]}}
\toprule
 &  & \multicolumn{2}{c}{indoor\_flying1} & \multicolumn{2}{c}{indoor\_flying2} & \multicolumn{2}{c}{indoor\_flying3} & \multicolumn{2}{c}{outdoor\_day1}\\
 \cmidrule(l{1mm}r{1mm}){3-4}
 \cmidrule(l{1mm}r{1mm}){5-6}
 \cmidrule(l{1mm}r{1mm}){7-8}
 \cmidrule(l{1mm}r{1mm}){9-10}
\multicolumn{2}{c}{$dt=1$} 
&\text{AEE $\downarrow$} & \text{\%Out $\downarrow$} & \text{AEE $\downarrow$} & \text{\%Out $\downarrow$} 
&\text{AEE $\downarrow$} & \text{\%Out $\downarrow$} & \text{AEE $\downarrow$} & \text{\%Out $\downarrow$}\\
\midrule 
\multirow{3}{*}{\begin{turn}{90}
SL
\end{turn}} 
 & EV-FlowNet-EST \cite{Gehrig19iccv} & 0.97 & 0.91 & 1.38 & 8.20 & 1.43 & 6.47 & {\textendash} & {\textendash} \\
 & EV-FlowNet+ \cite{Stoffregen20eccv} & 0.56 & 1.00 & \unum{1.2}{0.66} & \unum{1.2}{1.00} & \unum{1.2}{0.59} & 1.00 & 0.68 & 0.99\\
 & E-RAFT \cite{Gehrig21threedv} & {\textendash} & {\textendash} & {\textendash} & {\textendash} & {\textendash} & {\textendash} & \bnum{0.24} & 1.7\\\hline

\multirow{3}{*}{\begin{turn}{90}
SSL
\end{turn}}
 & EV-FlowNet (original) \cite{Zhu18rss} & 1.03 & 2.20 & 1.72 & 15.1 & 1.53 & 11.9 & 0.49 & 0.2\\
 & Spike-FlowNet \cite{Lee20eccv} & 0.84 &  {\textendash} & 1.28 & {\textendash} & 1.11 & {\textendash} & 0.49 & {\textendash} \\
 & Ziluo et al. \cite{Ziluo21arxiv} & 0.57 & 0.1 & 0.79 & 1.6 & 0.72 & 1.3 & 0.42 & \bnum{0.0}\\\hline

\multirow{4}{*}{\begin{turn}{90}
USL
\end{turn}}

 & EV-FlowNet \cite{Zhu19cvpr} & 0.58 & \bnum{0.0} & 1.02 & 4.0 & 0.87 & 3.0 & 0.32 & \bnum{0.0}\\
 
 & EV-FlowNet (retrained) \cite{Paredes21cvpr} & 0.79 & 1.2 & 1.40 & 10.9 & 1.18 & 7.4 & 0.92 & 5.4\\
 
 & FireFlowNet \cite{Paredes21cvpr} & 0.97 & 2.6 & 1.67 & 15.3 & 1.43 & 11.00 & 1.06 & 6.6\\
 
 & ConvGRU-EV-FlowNet \cite{Paredes21neurips} & 0.60 & 0.51 & 1.17 & 8.06 & 0.93 & 5.64 & 0.47 & 0.25\\\hline

\multirow{6}{*}{\begin{turn}{90}
MB
\end{turn}}
 & Nagata et al. \cite{Nagata21sensors} & 0.62 & {\textendash}  & 0.9324 & {\textendash} & 0.8436 & {\textendash} & 0.7696 & {\textendash} \\
 
 & Akolkar et al. \cite{Akolkar22pami} & 1.52 & {\textendash} & 1.59 & {\textendash} & 1.89 & {\textendash} & 2.75 &  {\textendash} \\
 
 & Brebion et al. \cite{Brebion21tits} & \unum{1.2}{0.52} & 0.1 & 0.98 & 5.5 & 0.71 & 2.1 & 0.53 & 0.2\\

 & Ours (w/o time aware) & \bnum{0.42152} & \unum{1.2}{0.09331} & \bnum{0.60318} & \bnum{0.59478} & \bnum{0.49978} & \unum{1.2}{0.28883} & \unum{1.2}{0.30041} & 0.10531 \\

 & Ours (Upwind) & \bnum{0.42281} & 0.10210 & \bnum{0.60418} & \bnum{0.58549} & \bnum{0.50099} & \bnum{0.28177} & \unum{1.2}{0.300529} & \unum{1.2}{0.10381} \\

 & Ours (Burgers') & \bnum{0.42274} & 0.10303 & \bnum{0.60456} & \bnum{0.59485} & \bnum{0.49987} & \bnum{0.27932} & \unum{1.2}{0.30075} & \unum{1.2}{0.10440} \\\hline
 \\[-0.2ex]
 
\multicolumn{2}{c}{$dt=4$} & & & & & & &  & \\
\midrule
\multirow{3}{*}{\begin{turn}{90}
SSL
\end{turn}}
 
 & EV-FlowNet (original) \cite{Zhu18rss} & 2.25 & 24.7 & 4.05 & 45.3 & 3.45 & 39.7 & 1.23 & \unum{1.2}{7.3}\\
 
 & Spike-FlowNet \cite{Lee20eccv} & 2.24 & {\textendash} & 3.83 & {\textendash} & 3.18 & {\textendash} & \unum{1.2}{1.09} & {\textendash} \\
 
 & Ziluo et al. \cite{Ziluo21arxiv} & 1.77 & 14.7 & \unum{1.2}{2.52} & \bnum{26.1} & \unum{1.2}{2.23} & 22.1 & \bnum{0.99} & \bnum{3.9}\\\hline

\multirow{2}{*}{\begin{turn}{90}
USL
\end{turn}}
 
 & EV-FlowNet \cite{Zhu19cvpr}&  2.18 & 24.2 & 3.85 & 46.8 & 3.18 & 47.8 & 1.30 & 9.7\\
 
 & ConvGRU-EV-FlowNet \cite{Paredes21neurips} & 2.16 & 21.51 & 3.90 & 40.72 & 3.00 & 29.60 & 1.69 & 12.50\\\hline

\multirow{3}{*}{\begin{turn}{90}
MB
\end{turn}}
 & Ours (w/o time aware) & \bnum{1.68217} & \bnum{12.79074} & \bnum{2.48917} & \unum{2.2}{26.31331} & \bnum{2.05633} & \bnum{18.92672} & 1.24740 & 9.19271 \\
 
 & Ours (Upwind) & \unum{1.2}{1.68777} & \unum{2.2}{12.83141} & \bnum{2.49070} & 26.37168 & \bnum{2.06188} & \unum{2.2}{19.02332} & 1.25065 & 9.227 \\
 
 & Ours (Burgers') & \unum{1.2}{1.68961} & 12.94904 & \bnum{2.48956} & 26.34562 & \bnum{2.06367} & 19.02718 & 1.24994 &  9.20702 \\
\bottomrule
\end{tabular}
}
\end{table}

%% file: floats/fig_qualitative_large.tex
\def\figWidth{0.192\linewidth} \def\figSquareWidth{0.144277457\linewidth} %
 \begin{figure*}[h!]
	\centering
    {\scriptsize
    \setlength{\tabcolsep}{1pt}
	\begin{tabular}{
	>{\centering\arraybackslash}m{\figWidth} 
	>{\centering\arraybackslash}m{\figWidth}
	>{\centering\arraybackslash}m{\figWidth}
	>{\centering\arraybackslash}m{\figSquareWidth}
	>{\centering\arraybackslash}m{\figSquareWidth}}
		\\%\addlinespace[1ex]

		{\makecell{indoor\_flying1}}
		&\gframe{\includegraphics[width=\linewidth]{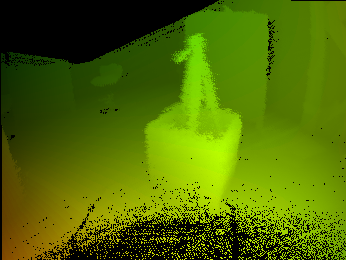}}
		&\gframe{\includegraphics[width=\linewidth]{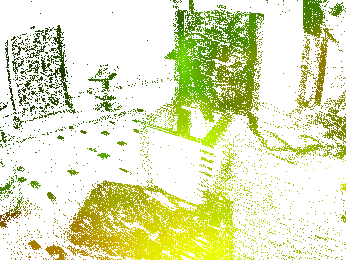}}
		&\gframe{\includegraphics[width=\linewidth]{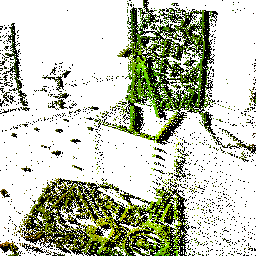}}
		&\gframe{\includegraphics[width=\linewidth]{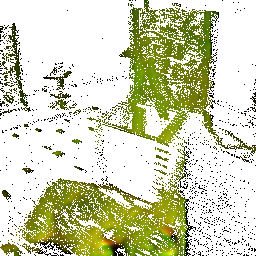}}
		\\
		\gframe{\includegraphics[width=\linewidth]{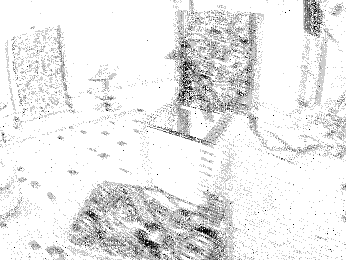}}
		&\gframe{\includegraphics[width=\linewidth]{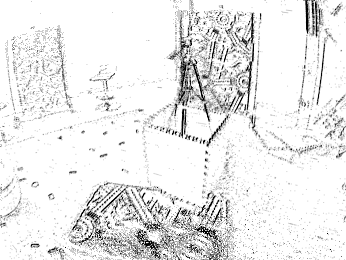}}
		&\gframe{\includegraphics[width=\linewidth]{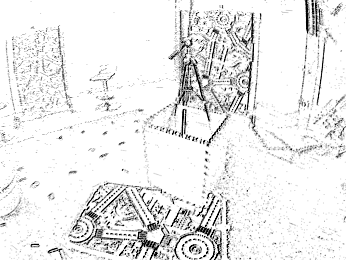}}
		&\gframe{\includegraphics[width=\linewidth]{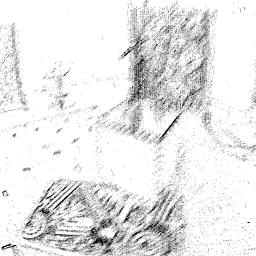}}
		&\gframe{\includegraphics[width=\linewidth]{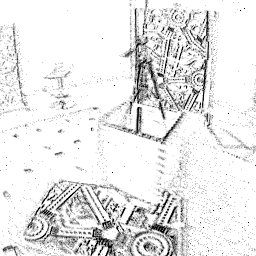}}
		\\

		{\makecell{indoor\_flying2}}
		&\gframe{\includegraphics[width=\linewidth]{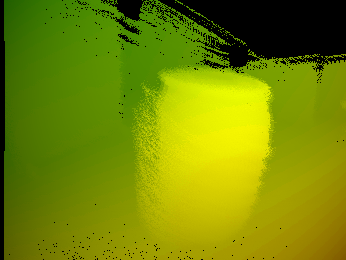}}
		&\gframe{\includegraphics[width=\linewidth]{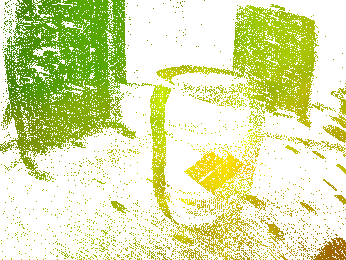}}
		&\gframe{\includegraphics[width=\linewidth]{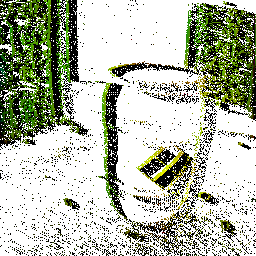}}
		&\gframe{\includegraphics[width=\linewidth]{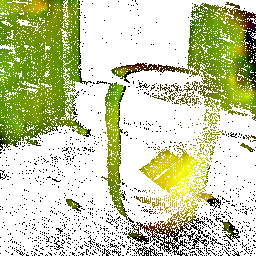}}
		\\
		\gframe{\includegraphics[width=\linewidth]{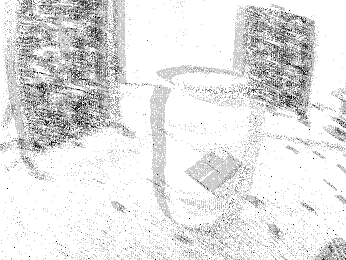}}
		&\gframe{\includegraphics[width=\linewidth]{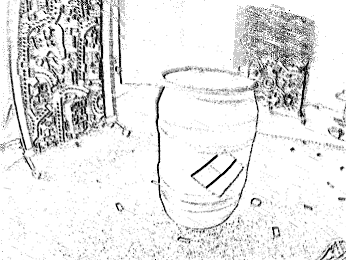}}
		&\gframe{\includegraphics[width=\linewidth]{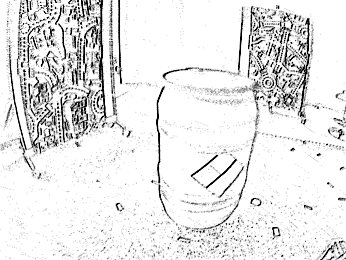}}
		&\gframe{\includegraphics[width=\linewidth]{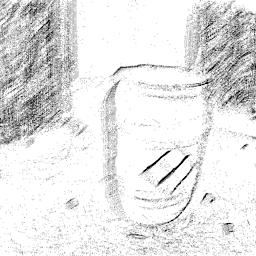}}
		&\gframe{\includegraphics[width=\linewidth]{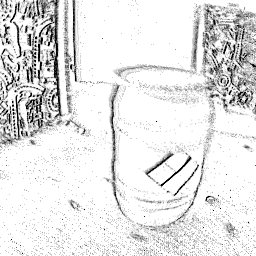}}
		\\

		{\makecell{indoor\_flying3}}
		&\gframe{\includegraphics[width=\linewidth]{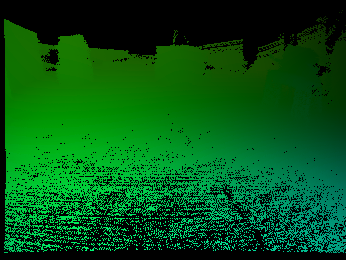}}
		&\gframe{\includegraphics[width=\linewidth]{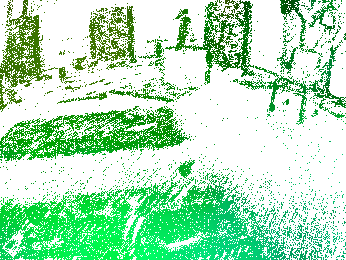}}
		&\gframe{\includegraphics[width=\linewidth]{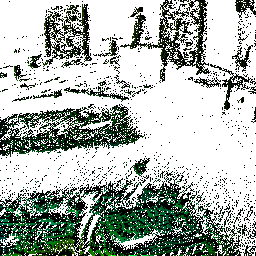}}
		&\gframe{\includegraphics[width=\linewidth]{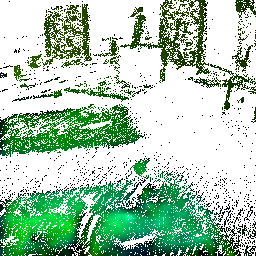}}
		\\
		\gframe{\includegraphics[width=\linewidth]{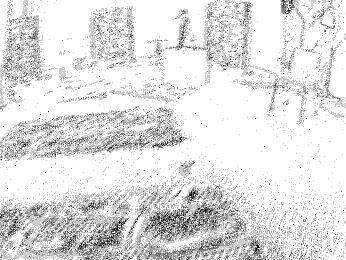}}
		&\gframe{\includegraphics[width=\linewidth]{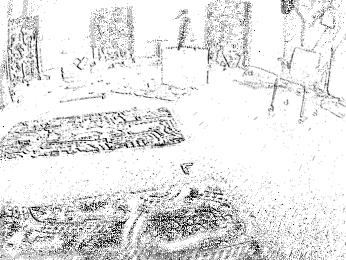}}
		&\gframe{\includegraphics[width=\linewidth]{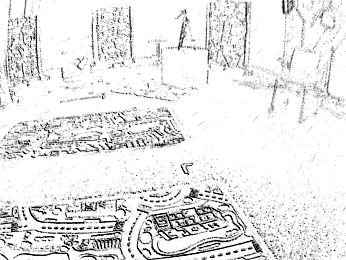}}
		&\gframe{\includegraphics[width=\linewidth]{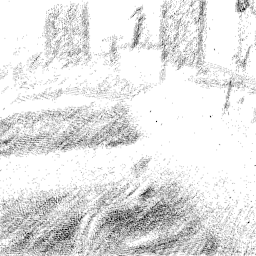}}
		&\gframe{\includegraphics[width=\linewidth]{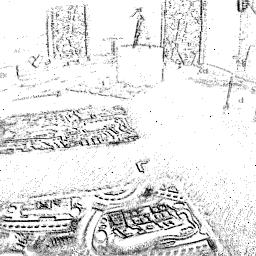}}
		\\

		{\makecell{outdoor\_day1}}
        &\gframe{\includegraphics[width=\linewidth]{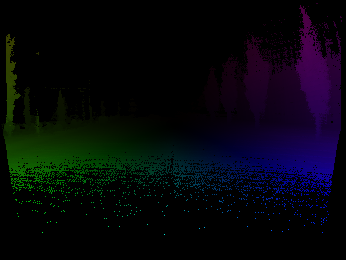}}
		&\gframe{\includegraphics[width=\linewidth]{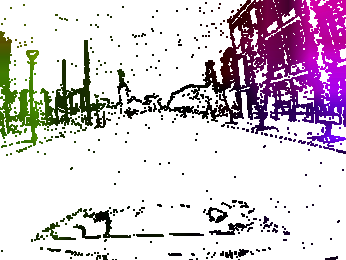}}
		&\gframe{\includegraphics[width=\linewidth]{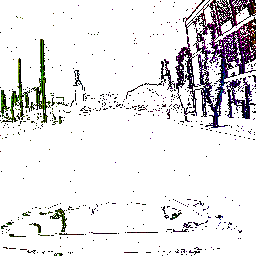}}
		&\gframe{\includegraphics[width=\linewidth]{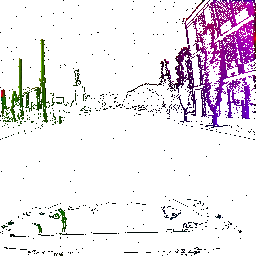}}
		\\
		\gframe{\includegraphics[width=\linewidth]{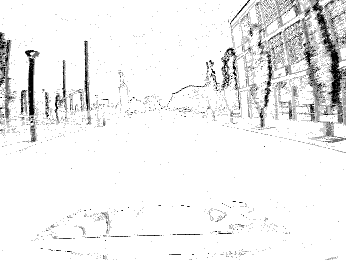}}
		&\gframe{\includegraphics[width=\linewidth]{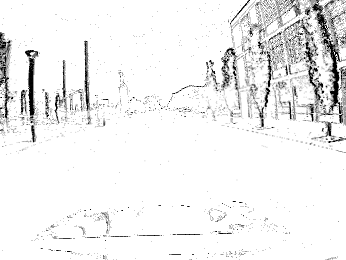}}
		&\gframe{\includegraphics[width=\linewidth]{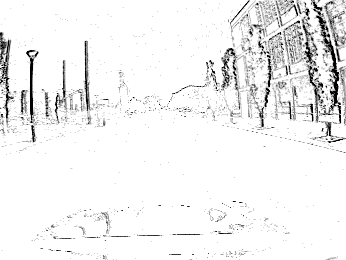}}
		&\gframe{\includegraphics[width=\linewidth]{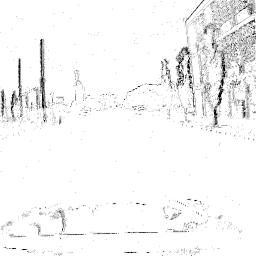}}
		&\gframe{\includegraphics[width=\linewidth]{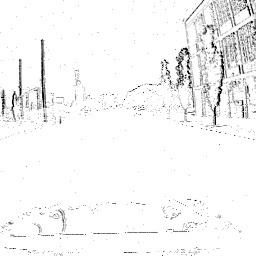}}
		\\

		(a) Events
		& (b) GT
		& (c) Ours (MB)
		& (d) USL\!\cite{Paredes21neurips}\!
		& (e) SSL\!\cite{Zhu18rss}
		\\%\addlinespace[1ex]
	\end{tabular}
	}
	\caption{\emph{MVSEC comparison} ($dt=4$) of our method and two state-of-the-art baselines:
	ConvGRU-EV-FlowNet (USL) \cite{Paredes21neurips}
	and EV-FlowNet (SSL) \cite{Zhu18rss}.
	For each sequence, the upper row shows the flow masked by the input events, and the lower row shows the IWE using the flow.
	Our method produces the sharpest motion-compensated IWEs. 
	Note that learning-based methods crop input events to center 256 $\times$ 256 pixels, whereas our method does not.
	Black points in ground truth (GT) flow maps indicate the absence of LiDAR measurements.
	The optical flow color wheel is in Fig. \ref{fig:eye_catcher}.
	}
	\label{fig:compare}
\label{fig:qualitative_comparison}
\end{figure*}

%% file: floats/tab_dsec.tex
\begin{table}[t]
\centering
\caption{Results on DSEC test sequences \cite{Gehrig21threedv}. 
For the calculation of FWL, we use events within 100ms. 
More sequences are provided in the supplementary material.
}
\label{tab:dsec}
\adjustbox{max width=\textwidth}{%
\setlength{\tabcolsep}{3pt}
\begin{tabular}{l*{9}{S[table-format=2.4]}}
\toprule
   & \multicolumn{3}{c}{thun\_01\_a}
  & \multicolumn{3}{c}{thun\_01\_b}
  & \multicolumn{3}{c}{zurich\_city\_15\_a} \\
 \cmidrule(l{1mm}r{1mm}){2-4}
 \cmidrule(l{1mm}r{1mm}){5-7}
 \cmidrule(l{1mm}r{1mm}){8-10}
 & \text{AEE $\downarrow$} & \text{\%Out $\downarrow$} & \text{FWL $\uparrow$}
 & \text{AEE $\downarrow$} & \text{\%Out $\downarrow$} & \text{FWL $\uparrow$} 
 & \text{AEE $\downarrow$} & \text{\%Out $\downarrow$} & \text{FWL $\uparrow$}  \\
\midrule

 E-RAFT \cite{Gehrig21threedv}
 & \bnum{0.654} & \bnum{1.87} & 1.200584411 
 & \bnum{0.578} & \bnum{1.518} & 1.1767931182449056
 & \bnum{0.589} & \bnum{1.303} & 1.335840906676725 \\ 
 Ours
 & 2.116 & 17.684 & \bnum{1.2420968732859203}
 & 2.48 & 23.564 & \bnum{1.241943689019936}
 & 2.347 & 20.987 & \bnum{1.4119389239425226} \\

\bottomrule
\end{tabular}
}
\end{table}

%% file: floats/fig_qualitative_dsec.tex
\global\long\def\figWidth{0.195\linewidth}
 \begin{figure*}[t]
	\centering
    {\scriptsize
    \setlength{\tabcolsep}{1pt}
	\begin{tabular}{
	>{\centering\arraybackslash}m{\figWidth} 
	>{\centering\arraybackslash}m{\figWidth} 
	>{\centering\arraybackslash}m{\figWidth}
	>{\centering\arraybackslash}m{\figWidth}
	>{\centering\arraybackslash}m{\figWidth}
	}
	    \gframe{\includegraphics[width=\linewidth]{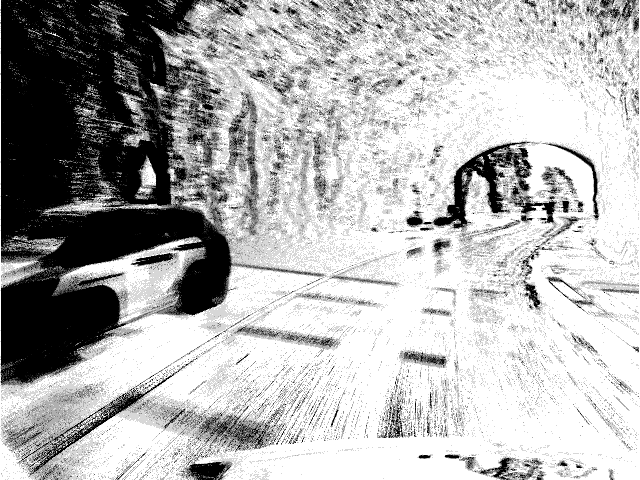}}
	    &\gframe{\includegraphics[width=\linewidth]{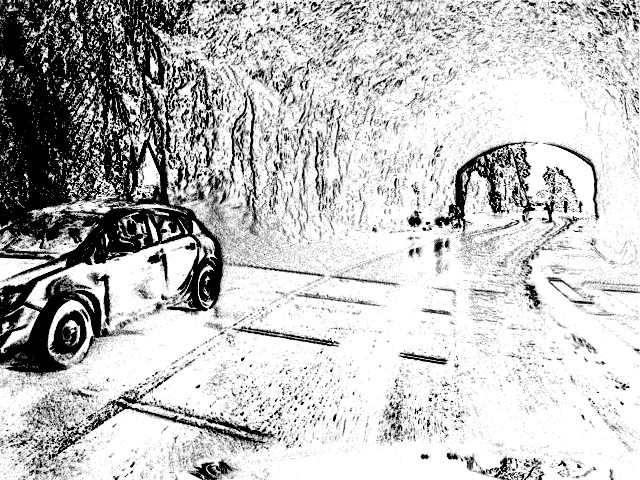}}
        &\gframe{\includegraphics[width=\linewidth]{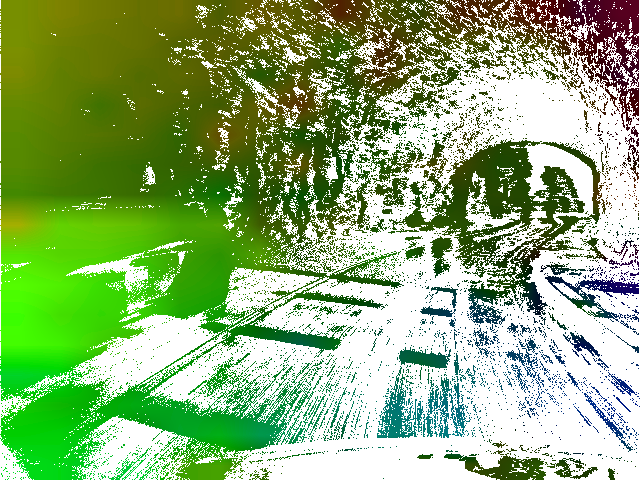}}
        &\gframe{\includegraphics[width=\linewidth]{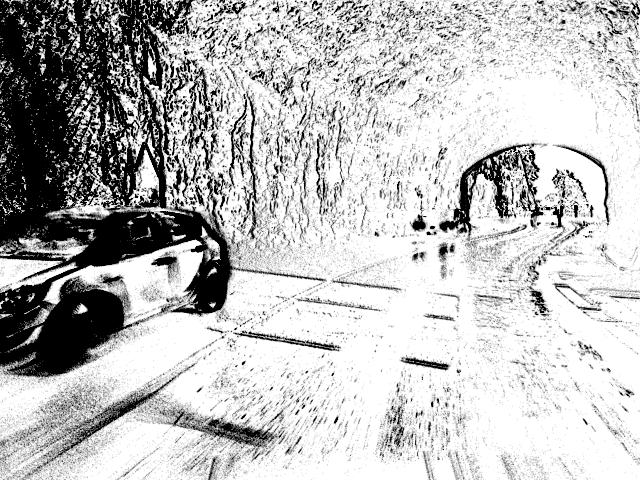}}
		&\gframe{\includegraphics[width=\linewidth]{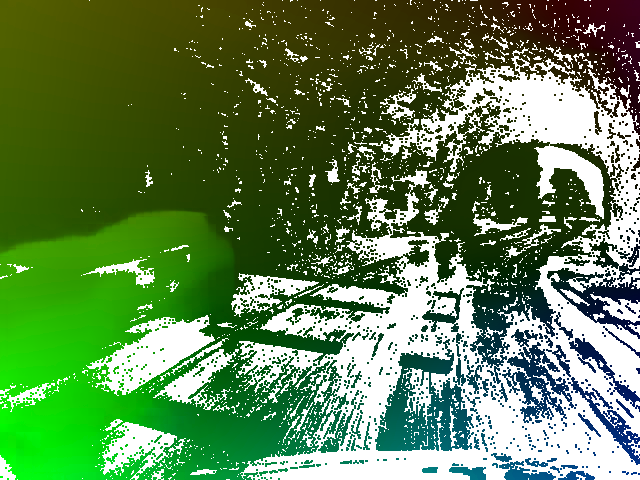}}
		\\

		(a) Events
		& (b) IWE (Ours)
		& (c) Flow (Ours)
		& (d) %
		IWE (SL) \cite{Gehrig21threedv}
		& (e) %
		Flow (SL) \cite{Gehrig21threedv}
		\\%\addlinespace[1ex]
	\end{tabular}
	}
	\caption{\emph{DSEC results} %
	on the interlaken\_00\_b test sequence (no GT available).
	Since GT is missing at IMOs and points outside the LiDAR's FOV,
	the supervised method \cite{Gehrig21threedv} may provide inaccurate predictions around IMOs and road points close to the camera, whereas our method produces sharp edges. 
	For visualization, we use 1M events.
	}
\label{fig:qualitative_dsec}
\end{figure*}

%% file: floats/fig_suppl_abl_multiref.tex
\def\figWidth{0.17\linewidth}
\begin{figure*}[t]
	\centering
    {\scriptsize
    \setlength{\tabcolsep}{1pt}
	\begin{tabular}{
	>{\centering\arraybackslash}m{0.3cm}
	>{\centering\arraybackslash}m{\figWidth} 
	>{\centering\arraybackslash}m{\figWidth}
	>{\centering\arraybackslash}m{\figWidth}
	>{\centering\arraybackslash}m{\figWidth}
	>{\centering\arraybackslash}m{\figWidth}}
		\\%\addlinespace[1ex]

		\rotatebox{90}{\makecell{Single ref. ($t_1$)}}
		&\gframe{\includegraphics[trim={200px 135px 0 0},clip,width=\linewidth]{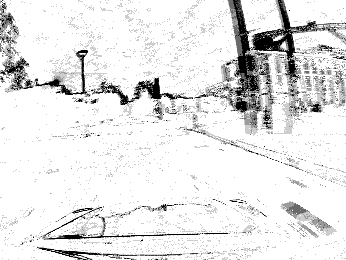}}
		&\gframe{\includegraphics[trim={200px 135px 0 0},clip,width=\linewidth]{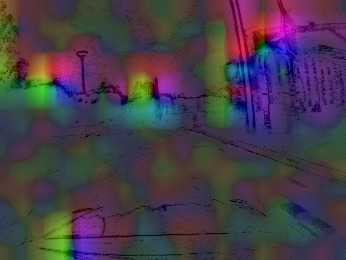}}
		&\gframe{\includegraphics[trim={200px 135px 0 0},clip,width=\linewidth]{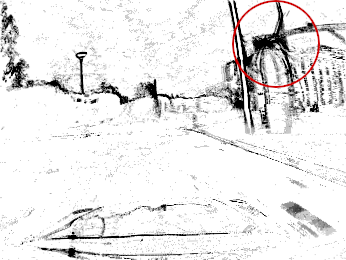}}
		&\gframe{\includegraphics[trim={200px 135px 0 0},clip,width=\linewidth]{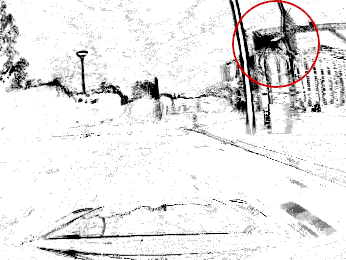}}
		&\gframe{\includegraphics[trim={200px 135px 0 0},clip,width=\linewidth]{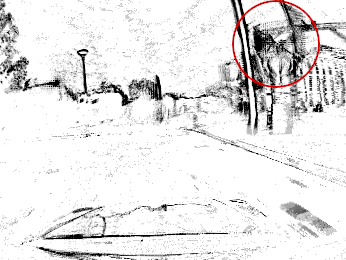}}
		\\
		\rotatebox{90}{\makecell{Multi-ref.}}
		& \gframe{\includegraphics[trim={200px 135px 0 0},clip,width=\linewidth]{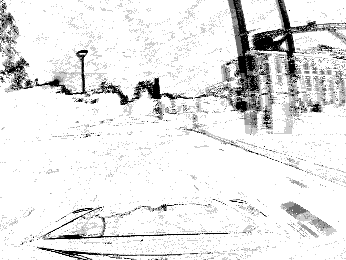}}
		&\gframe{\includegraphics[trim={200px 135px 0 0},clip,width=\linewidth]{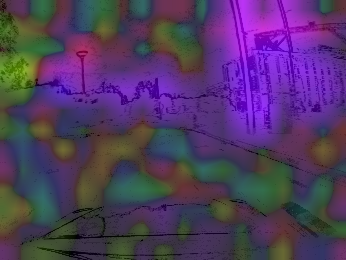}}
		&\gframe{\includegraphics[trim={200px 135px 0 0},clip,width=\linewidth]{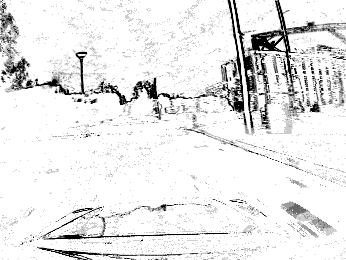}}
		&\gframe{\includegraphics[trim={200px 135px 0 0},clip,width=\linewidth]{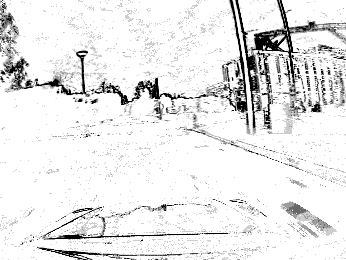}}
		&\gframe{\includegraphics[trim={200px 135px 0 0},clip,width=\linewidth]{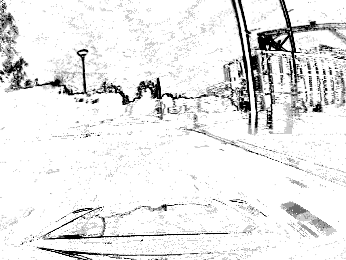}}
		\\

		& (a) Input events
		& (b) Flow
		& (c) Warp $t_1$
		& (d) Warp $\tmid$
		& (e) Warp $t_{\numEvents}$
		\\%\addlinespace[1ex]
	\end{tabular}
	}
	\caption{\emph{Effect of the multi-reference focus loss}. 
	}
\label{fig:suppl_abl_multiref}
\end{figure*}

%% file: floats/tab_fwl.tex
\begin{table}[t]
\centering
\caption{FWL (IWE sharpness) results on MVSEC, DSEC, and ECD. Higher is better.}
\label{tab:fwl}
\adjustbox{max width=\textwidth}{%
\setlength{\tabcolsep}{3pt}
\begin{tabular}{l*{7}{S[table-format=2.4]}}
\toprule
 & \multicolumn{4}{c}{\text{MVSEC ($dt=4$)}} & \text{ECD} & \multicolumn{2}{c}{\text{DSEC}}\\
 \cmidrule(l{1mm}r{1mm}){2-5}
 \cmidrule(l{1mm}r{1mm}){6-6}
 \cmidrule(l{1mm}r{1mm}){7-8}
 & \text{indoor1} & \text{indoor2} & \text{indoor3} & \text{outdoor1} & \text{slider\_depth} & \text{thun\_00\_a} & \text{zurich\_city\_07\_a}\\
\midrule 
Ground truth & 1.09 & 1.20 & 1.12 & 1.07 & {\textendash} & 1.01 & 1.042 \\
Ours: w/o time aware & \bnum{1.17} & \bnum{1.30} & \bnum{1.23} & \bnum{1.11} & 1.882863 & 1.39 & 1.57\\
Ours: Upwind & \bnum{1.17} & \bnum{1.30} & \bnum{1.23} & \bnum{1.1112} & 1.9246846078539779 & 1.40333 & 1.5951691658176472 \\
Ours: Burgers' & \bnum{1.17} & \bnum{1.30} & \bnum{1.23} & \bnum{1.11} & \bnum{1.9298641427098424} & \bnum{1.42} & \bnum{1.63}\\
\bottomrule
\end{tabular}
}
\end{table}

%% file: floats/fig_effect_of_time_awareness.tex
\def\figWidth{0.185\linewidth}
\begin{figure*}[t]
	\centering
    {\scriptsize
    \setlength{\tabcolsep}{1pt}
	\begin{tabular}{
	>{\centering\arraybackslash}m{0.3cm}
	>{\centering\arraybackslash}m{\figWidth} 
	>{\centering\arraybackslash}m{\figWidth}
	>{\centering\arraybackslash}m{\figWidth}
	>{\centering\arraybackslash}m{\figWidth}
	>{\centering\arraybackslash}m{\figWidth}}
		\\%\addlinespace[1ex]

		\rotatebox{90}{\makecell{slider\_depth}}
		&\gframe{\includegraphics[trim={0 90px 120px 0},clip,width=\linewidth]{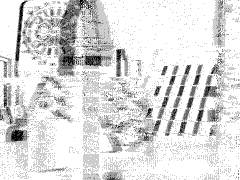}}
		&\gframe{\includegraphics[trim={0 90px 120px 0},clip,width=\linewidth]{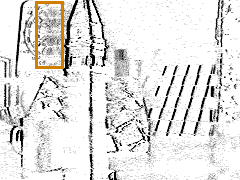}}
		&\gframe{\includegraphics[trim={0 90px 120px 0},clip,width=\linewidth]{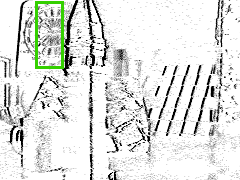}}
		&\gframe{\includegraphics[trim={0 90px 120px 0},clip,width=\linewidth]{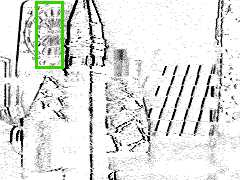}}
		& {\makecell{N.A.}}
		\\%\addlinespace[1ex]
		
		\rotatebox{90}{\!\!\!\!\makecell{zurich\_city\_11c}}

		&\gframe{\includegraphics[trim={0 150px 450px 170px},clip,width=\linewidth]{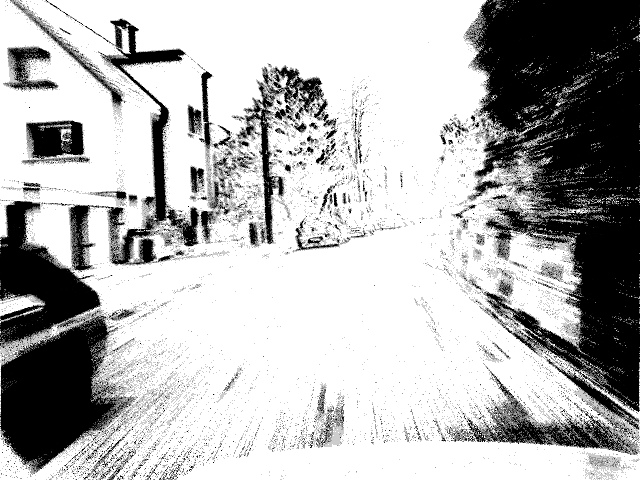}}
        &\gframe{\includegraphics[trim={0 150px 450px 170px},clip,width=\linewidth]{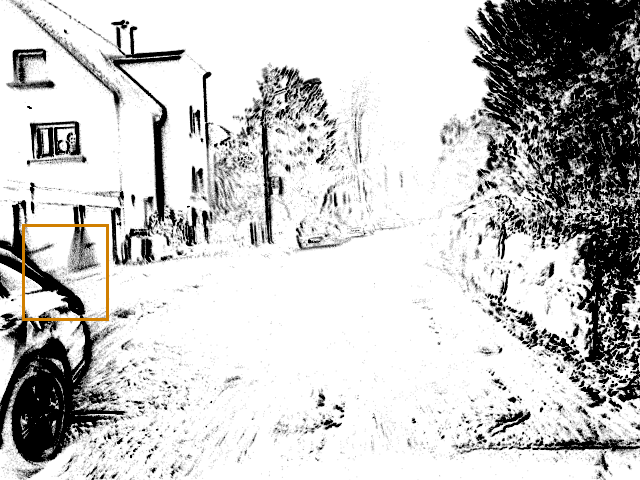}}
        &\gframe{\includegraphics[trim={0 150px 450px 170px},clip,width=\linewidth]{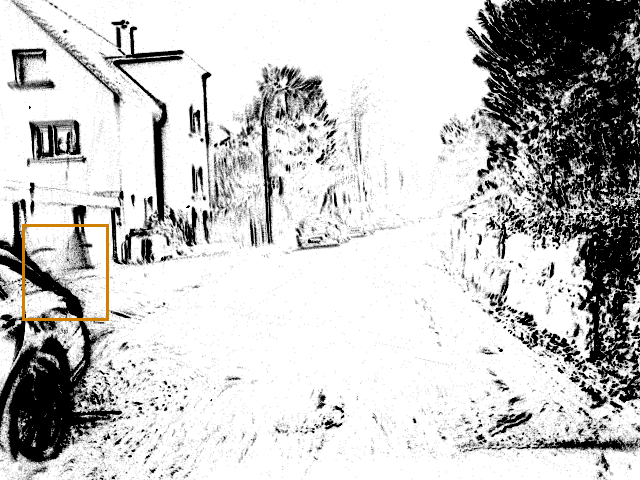}}
		&\gframe{\includegraphics[trim={0 150px 450px 170px},clip,width=\linewidth]{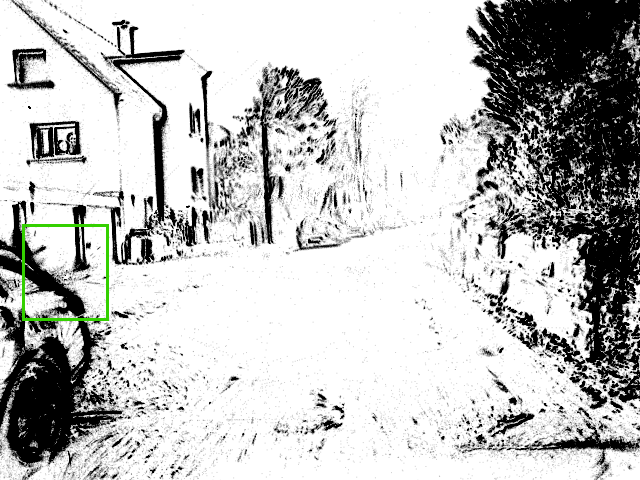}}
		&\gframe{\includegraphics[trim={0 150px 450px 170px},clip,width=\linewidth]{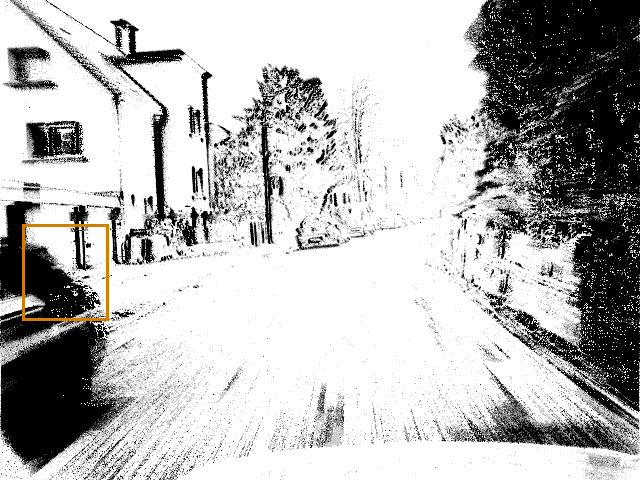}}
		\\%\addlinespace[1ex]

		& (a) Events
		& (b) Warp \eqref{eq:warp:oflow} %
		& (c) \eqref{eq:warp:oflow:PDE}-Upwind
		& (d) \eqref{eq:warp:oflow:PDE}-Burgers'
		& (e) Warp with GT
		\\%\addlinespace[1ex]
	\end{tabular}
	}
	\caption{\emph{Time-aware flow}.
	Comparison between 3 versions of our method: 
	Burgers', upwind, and no time-aware \eqref{eq:warp:oflow}.
	At occlusions (dartboard in slider\_depth \cite{Mueggler17ijrr} and garage door in DSEC \cite{Gehrig21ral}), upwind and Burgers' produce sharper IWEs.
	Due to the smoothness of the flow conferred by the tile-based approach, some small regions are still blurry.
	}
\label{fig:effect_of_time_aware}
\end{figure*}

%% file: floats/tab_dnn.tex
\begin{table}[t]
\centering
\caption{Results of unsupervised learning on MVSEC's outdoor\_day1 sequence.
}
\label{tab:dnn}
\adjustbox{max width=0.9\textwidth}{%
\setlength{\tabcolsep}{3pt}
\begin{tabular}{l*{7}{S[table-format=2.3]}}
\toprule
 & \multicolumn{3}{c}{$dt=1$} & &\multicolumn{3}{c}{$dt=4$} \\
 \cmidrule(l{1mm}r{1mm}){2-4}
 \cmidrule(l{1mm}r{1mm}){6-8}
 & \text{AEE $\downarrow$} & \text{\%Out $\downarrow$} & \text{FWL $\uparrow$} &
 & \text{AEE $\downarrow$} & \text{\%Out $\downarrow$} & \text{FWL $\uparrow$}\\
\midrule
 EV-FlowNet \cite{Zhu19cvpr} & \bnum{0.32} & \bnum{0.0} &  {\textendash} & & \bnum{1.30} & \bnum{9.7} & {\textendash}\\
 EV-FlowNet (retrained) \cite{Paredes21cvpr} & 0.92 & 5.4 & {\textendash} & & {\textendash} & {\textendash} & {\textendash} \\
 ConvGRU-EV-FlowNet \cite{Paredes21neurips} & 0.47 & 0.25 & 0.94244531  & & 1.69 & 12.50 & 0.94381884886 \\
 Our EV-FlowNet using \eqref{eq:composite:function} & \unum{1.2}{0.3632} & \unum{1.2}{0.087} & \bnum{0.9622}  & & \unum{1.2}{1.488} & \unum{2.2}{11.72} & \bnum{1.114} \\
\bottomrule
\end{tabular}
}
\end{table}

%% file: chapters/05_limitations.tex
\subsection{Limitations}
\label{sec:limitations}
Like previous unsupervised works \cite{Zhu19cvpr,Paredes21neurips}, our method is based on the brightness constancy assumption. 
Hence, it struggles to estimate flow from events that are not due to motion, such as those caused by flickering lights. %
SL and SSL methods may forego this assumption, but they require high quality supervisory signal, which is challenging due to the HDR and high speed of event cameras.

Like other optical flow methods, our approach can suffer from the aperture problem.
The flow could still collapse (events may be warped to too few pixels) 
if tiles become smaller (higher DOFs), or without proper regularization or initialization.
Optical flow is also difficult to estimate in regions with few events, 
such as homogeneous brightness regions and regions with small apparent motion.
Regularization fills in the homogeneous regions, 
whereas recurrent connections (like in RNNs) could help with small apparent motion.

%% file: chapters/06_conclusion.tex
\section{Conclusion}
\label{sec:conclusion}

We have extended the CM framework to estimate dense optical flow,
proposing principled solutions to overcome problems of overfitting, occlusions and convergence without performing event voxelization. 
The comprehensive experiments show that our method achieves the best accuracy among all methods in the MVSEC indoor benchmark,
and among the unsupervised and model-based methods in the outdoor sequence.
It is also competitive in the DSEC optical flow benchmark.
Moreover, our method delivers the sharpest IWEs and exposes the limitations of the benchmark data.
Finally, we show how our method can be ported to the unsupervised setting, producing remarkable results.
We hope our work unlocks future optical flow research on stable and interpretable methods.

%% file: chapters/06_ack.tex
\textbf{Acknowledgements}.
We thank Prof. A. Yezzi and Dr. A. Zhu %
for useful discussions.
Funded by the German Academic Exchange Service (DAAD), Research Grant - Bi-nationally Supervised Doctoral Degrees/Cotutelle, 2021/22 (57552338).
Funded by the Deutsche Forschungsgemeinschaft (DFG, German Research Foundation) under Germany’s Excellence Strategy – EXC 2002/1 ``Science of Intelligence'' – project number 390523135.

%% file: chapters/07_suppl_mat.tex
\setcounter{secnumdepth}{3}

\appendix

\section{Supplementary Material}

\subsection{Time-Awareness: PDE solutions}

The proposed \emph{time-aware flow} is given as the solution to \eqref{eq:flow:pde}.
Letting the flow be $\velflow = (v_{x}, v_{y})^\top$, the system of PDEs can be written as:
\begin{equation}
\begin{split}
v_{x} \prtl{v_x}{x} + v_{y} \prtl{v_x}{y} + \prtl{v_x}{t} = 0,\\
v_{x} \prtl{v_y}{x} + v_{y} \prtl{v_y}{y} + \prtl{v_y}{t} = 0.
\end{split}
\end{equation}
Upwind and Burgers' schemes can be used to discretize and numerically solve the system of PDEs \cite{Evans10book,Sethian99book}.

\textbf{Discretization}. 
Let $\velflow^{n}(x, y)$ be the flow vector at discretized space- (e.g., pixel) and time-indices $(x, y, n)$, 
with discretization steps $\Delta x, \Delta y$, and $\Delta t$, respectively, 
and let the forward ($+$) and backward ($-$) differences of a scalar field $w$ (e.g., $v^{n}_{x}$ or $v^{n}_{y}$) be defined as
\begin{equation}
\begin{split}
\label{eq:suppl:fwddiff}
D_x^+ w \equiv \prtl{w}{x}^{+} = \frac1{\Delta x}\bigl(w(x + \Delta x, y) - w(x,y)\bigr),\\
D_y^+ w \equiv \prtl{w}{y}^{+} = \frac1{\Delta y}\bigl(w(x, y + \Delta y) - w(x,y)\bigr),
\end{split}
\end{equation}
and
\begin{equation}
\begin{split}
\label{eq:suppl:bwddiff}
D_x^- w \equiv \prtl{w}{x}^{-} = \frac1{\Delta x}\bigl(w(x,y) - w(x - \Delta x, y)\bigr),\\
D_y^- w \equiv \prtl{w}{y}^{-} = \frac1{\Delta y}\bigl(w(x,y) - w(x, y - \Delta y)\bigr).
\end{split}
\end{equation}
We discretize in time using forward differences, $\prtl{w}{t} \approx (w(t+\Delta t)-w(t)) / \Delta t$, 
to yield explicit update schemes: 
$w(t+\Delta t) \approx w(t) + \Delta t \prtl{w}{t}$.

\textbf{Upwind scheme}. 
The first-order upwind scheme is an explicit scheme that updates the flow as follows,
based on the sign of the variables:
it uses $\fwd{x}{v^{n}_{x}}$ and $\fwd{x}{v^{n}_{y}}$ for $v^{n}_{x} > 0$ ($\bwd{x}{v^{n}_{x}}$ and $\bwd{x}{v^{n}_{y}}$ otherwise), 
and $\fwd{y}{v^{n}_{x}}$ and $\fwd{y}{v^{n}_{y}}$ for $v^{n}_{y} > 0$ ($\bwd{y}{v^{n}_{x}}$ and $\bwd{y}{v^{n}_{y}}$ otherwise).
The scheme is stable if the flow satisfies 
$\Delta t \max\{{|v_{x}|/\Delta x + |v_{y}|/\Delta y}\} < 1$ (CFL stability condition \cite{Hirsch07book}).
For example, in case that $v^{n}_{x} > 0$ and $v^{n}_{y} > 0$ at the current discretization time $n$:
\begin{equation}
\begin{split}
\label{eq:suppl:upwind}
v^{n+1}_{x} = v^{n}_{x} - \Delta t \left(v^{n}_{x} \fwd{x}{v^{n}_{x}} + v^{n}_{y} \fwd{y}{v^{n}_{x}} \right),\\
v^{n+1}_{y} = v^{n}_{y} - \Delta t \left(v^{n}_{y} \fwd{y}{v^{n}_{y}} + v^{n}_{x} \fwd{x}{v^{n}_{y}} \right).
\end{split}
\end{equation}

\textbf{Burgers' scheme}. 
The study of the inviscid Burgers' equation provides a more conservative scheme solution, 
especially at ``shock'' and ``fan wave'' cases \cite{Sethian99book}.
In this explicit scheme, the product terms in the same variable (which convey that the flow is transporting itself), $v^{n}_{x} \fwd{x}{v^{n}_{x}}$ and $v^{n}_{y} \fwd{y}{v^{n}_{y}}$ in \eqref{eq:suppl:upwind}, are replaced with $\uxx$ and $\uyy$ respectively, which are given by:
\begin{equation}
\begin{split}
\label{eq:suppl:burgerstermx}
\uxx = \frac{1}{2} \biggl( \sign\bigl(v^{n}_{x}(x, y)\bigr) \bigl(v^{n}_{x}(x, y) \bigr)^2 + F_{x} - B_{x} \biggr),\\
F_{x} = 
\begin{cases}
    \bigl(v^{n}_{x}(x + \Delta x, y) \bigr)^2,& \text{if } v^{n}_{x}(x + \Delta x, y) < 0\\
    0,              & \text{otherwise}
\end{cases}\\
B_{x} = 
\begin{cases}
    \bigl(v^{n}_{x}(x - \Delta x, y)\bigr)^2,& \text{if } v^{n}_{x}(x - \Delta x, y) > 0\\
    0,              & \text{otherwise}
\end{cases}
\end{split}
\end{equation}
and
\begin{equation}
\begin{split}
\label{eq:suppl:burgerstermy}
\uyy = \frac{1}{2} \biggl( \sign\bigl(v^{n}_{y}(x, y)\bigr)\bigl(v^{n}_{y}(x, y)\bigr)^2 + F_{y} - B_{y} \biggr),\\
F_{y} = 
\begin{cases}
    \bigl(v^{n}_{y}(x, y + \Delta y)\bigr)^2,& \text{if } v^{n}_{y}(x, y + \Delta y) < 0\\
    0,              & \text{otherwise}
\end{cases}\\
B_{y} = 
\begin{cases}
    \bigl(v^{n}_{y}(x, y - \Delta y)\bigr)^2,& \text{if } v^{n}_{y}(x, y - \Delta y) > 0\\
    0.              & \text{otherwise}
\end{cases}
\end{split}
\end{equation}

\subsection{Effect of the Multi-scale Approach}

The effect of the proposed multi-scale approach (Fig.~\ref{fig:fig_method_multi_scale}) is shown in Fig.~\ref{fig:suppl_abl_multiscale}.
This experiment compares the results of using multi-scale approaches (in a coarse-to-fine fashion) versus using a single (finest) scale.
With a single scale, the optimizer gets stuck in a local extremal, yielding an irregular flow field (see the optical flow rows), which may produce a blurry IWE (e.g., outdoor\_day1 scene).
With three scales (finest tile and two downsampled ones), the flow becomes less irregular than with one single scale, but there may be regions with few events where the flow is difficult to estimate.
With five scales the flow becomes smoother, more coherent over the whole image domain, while still being able to produce sharp IWEs.

\input{floats/fig_suppl_scale}

\subsection{Sensitivity Analysis}
\label{sec:experim:sensitivity}

\subsubsection{The choice of loss function.}

Table \ref{tab:suppl_loss} shows the results on the MVSEC benchmark for different loss functions.
We compare the (squared) gradient magnitude, image variance, 
average timestamp \cite{Zhu19cvpr}, and normalized average timestamp \cite{Paredes21neurips}.
The gradient magnitude and image variance losses produce the best accuracy compared with the two average timestamp losses.
Quantitatively, the image variance loss gives competitive results with respect to the gradient magnitude.
However, for the reasons described in Sec.~\ref{sec:method:multiref}, and because the image variance sometimes overfits, we use gradient magnitude.
Both average timestamp losses are trapped in the global optima which pushes all events out of the image plane, 
hence, the provide very large errors (marked as ``$>99$'' in Tab.~\ref{tab:suppl_loss}).
This effect is visualized in Fig.~\ref{fig:suppl_loss}.

\emph{Remark}: 
Maximization of \eqref{eq:theloss} does not suffer from the problem mentioned in \cite{Paredes21neurips} that affects the average timestamp loss function, namely that the optimal flow warps all events outside the image so as to minimize the loss (undesired global optima shown in Fig.~\ref{fig:suppl_loss}d-\ref{fig:suppl_loss}e).
If most events were warped outside of the image, then \eqref{eq:theloss} %
would be smaller than the identity warp, which contradicts maximization.

\input{floats/tab_suppl_loss}
\input{floats/fig_suppl_loss}

\subsubsection{The regularizer weight.}

Table \ref{tab:suppl_reg_weight} shows the sensitivity analysis on the regularizer weight $\lambda$ in~\eqref{eq:composite:function}.
$\lambda=0.0025$ provides the best accuracy in the outdoor sequence, while $\lambda=0.025$ provides slightly better accuracy in the indoor sequences.
Comparing their accuracy differences, we use the former because it has a higher accuracy gain.

\input{floats/tab_suppl_reg_weight}

\subsection{Additional Results}

\subsubsection{Full results on DSEC test sequences.}

For completeness, Tab.~\ref{tab:suppl_dsec} reports the results on all DSEC test sequences. 
No GT flow is available for these sequences.
As mentioned in Sec.~\ref{sec:experim:dsec}, only one competing method is available at the time of publication, which is supervised learning.
Our method consistently produces better FWL (i.e., sharper IWEs) than the supervised learning method, which suffers from GT issues to produce sharp IWEs.
The sharpness differences are most significant in IMOs and on the road close to the vehicle (see Fig.~\ref{fig:qualitative_dsec}).
The FWL is computed using the same 100ms intervals used for the accuracy benchmark calculation.
Since the FWL is sensitive to the number of events, the previous convention is consistent with the benchmark.

\input{floats/tab_suppl_dsec}

\subsubsection{Qualitative results for DNN.}

Additional qualitative results of our unsupervised learning setting (Sec. \ref{sec:experim:applytodnn}) are shown in Fig. \ref{fig:suppl_extra_qualitative}.
We compare our method with the state-of-the-art unsupervised learning \cite{Paredes21neurips}.
Our results resemble the GT flow.
See Tab.~\ref{tab:dnn} for the quantitative result.

\input{floats/fig_suppl_extra_quality}

%% file: floats/fig_suppl_scale.tex
\def\figWidth{0.24\linewidth}
\begin{figure*}[h!]
	\centering
    {\scriptsize
    \setlength{\tabcolsep}{1pt}
	\begin{tabular}{
	>{\centering\arraybackslash}m{\figWidth} 
	>{\centering\arraybackslash}m{\figWidth}
	>{\centering\arraybackslash}m{\figWidth}
	>{\centering\arraybackslash}m{\figWidth}}
		\\%\addlinespace[1ex]

		{\makecell{indoor\_flying1}}
		&\gframe{\includegraphics[width=\linewidth]{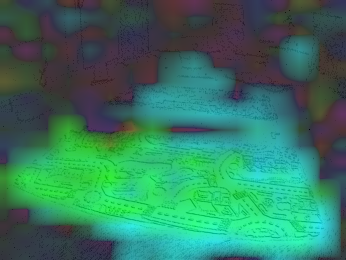}}
		&\gframe{\includegraphics[width=\linewidth]{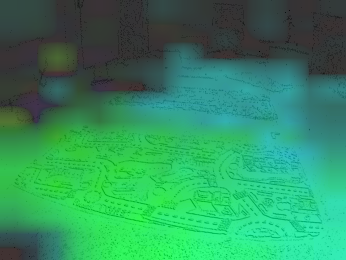}}
		&\gframe{\includegraphics[width=\linewidth]{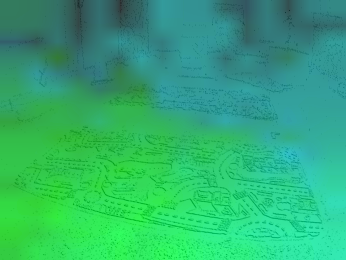}}
		\\
		&\gframe{\includegraphics[width=\linewidth]{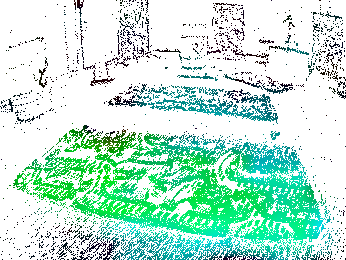}}
		&\gframe{\includegraphics[width=\linewidth]{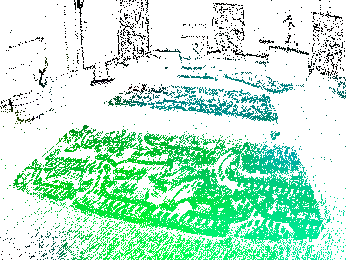}}
		&\gframe{\includegraphics[width=\linewidth]{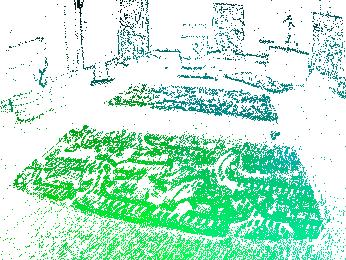}}
		\\
		\gframe{\includegraphics[width=\linewidth]{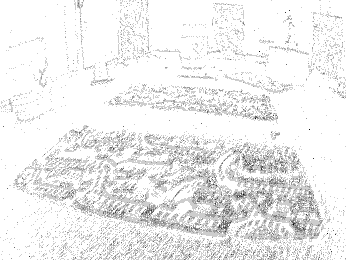}}
		&\gframe{\includegraphics[width=\linewidth]{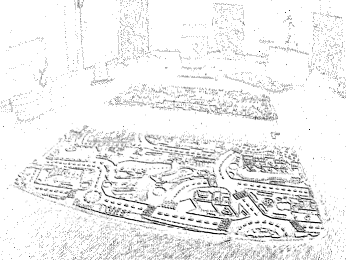}}
		&\gframe{\includegraphics[width=\linewidth]{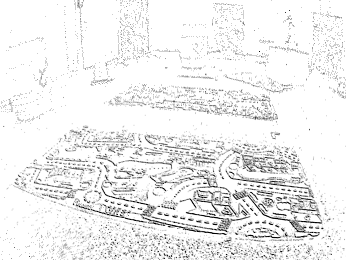}}
		&\gframe{\includegraphics[width=\linewidth]{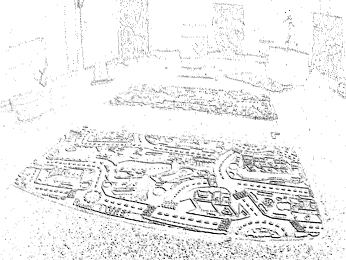}}
		\\

		{\makecell{outdoor\_day1}}
		&\gframe{\includegraphics[width=\linewidth]{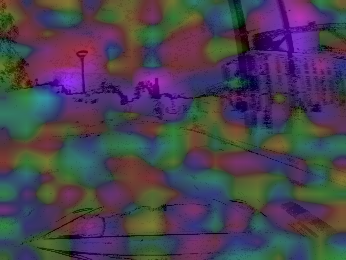}}
		&\gframe{\includegraphics[width=\linewidth]{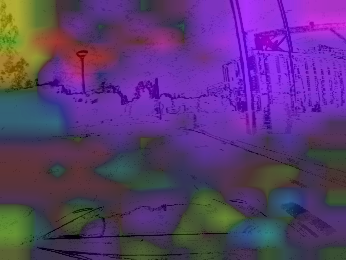}}
		&\gframe{\includegraphics[width=\linewidth]{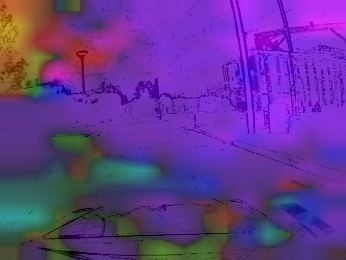}}
		\\
		&\gframe{\includegraphics[width=\linewidth]{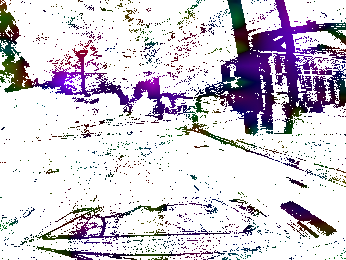}}
		&\gframe{\includegraphics[width=\linewidth]{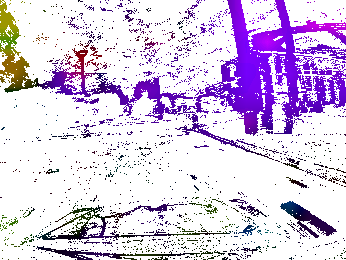}}
		&\gframe{\includegraphics[width=\linewidth]{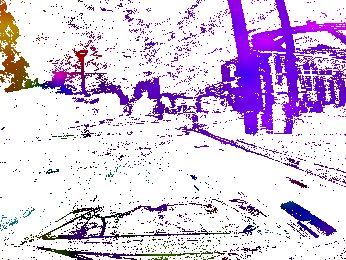}}
		\\
		\gframe{\includegraphics[width=\linewidth]{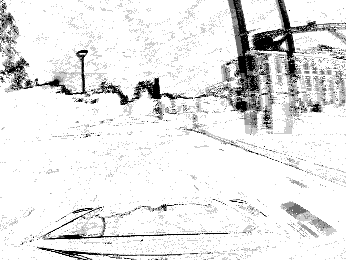}}
		&\gframe{\includegraphics[width=\linewidth]{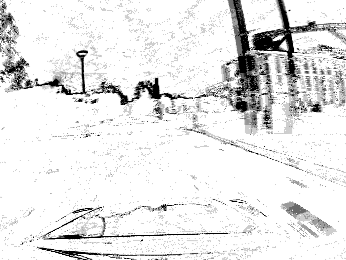}}
		&\gframe{\includegraphics[width=\linewidth]{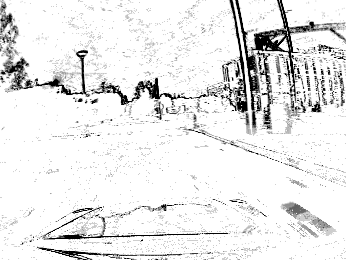}}
		&\gframe{\includegraphics[width=\linewidth]{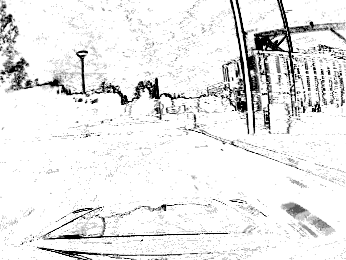}}
		\\

		(a) Input events
		& (b) Single (fine) scale
		& (c) Three scales %
		& (d) $N_{\ell}=5$ scales
		\\%\addlinespace[1ex]
	\end{tabular}
	}
	\caption{\emph{Effect of the multi-scale approach}.
	For each sequence, the top row shows the estimated flow, the middle row shows the estimated flow masked by the events, and the bottom row shows the IWEs.
	}
\label{fig:suppl_abl_multiscale}
\end{figure*}

%% file: floats/tab_suppl_loss.tex
\begin{table}[h!]
\centering
\caption{Sensitivity analysis on the choice of loss function (MVSEC, $dt=4$). 
The contrast and gradient magnitude functions provide notably better results than the losses based on average timestamps.
}
\label{tab:suppl_loss}
\adjustbox{max width=\textwidth}{%
\setlength{\tabcolsep}{2pt}
\begin{tabular}{l*{8}{S[table-format=2.2]}}
\toprule
  & \multicolumn{2}{c}{indoor\_flying1} 
  & \multicolumn{2}{c}{indoor\_flying2} 
  & \multicolumn{2}{c}{indoor\_flying3} 
  & \multicolumn{2}{c}{outdoor\_day1}\\
 \cmidrule(l{1mm}r{1mm}){2-3}
 \cmidrule(l{1mm}r{1mm}){4-5}
 \cmidrule(l{1mm}r{1mm}){6-7}
 \cmidrule(l{1mm}r{1mm}){8-9}
 &\text{AEE $\downarrow$} & \text{\%Out $\downarrow$} & \text{AEE $\downarrow$} & \text{\%Out $\downarrow$}  
 &\text{AEE $\downarrow$} & \text{\%Out $\downarrow$} & \text{AEE $\downarrow$} & \text{\%Out $\downarrow$}   
 \\
\midrule

 Gradient magnitude \cite{Gallego19cvpr} & \bnum{1.68217} & 12.79074 & 2.48917 & 26.31331 & 2.05633 & 18.92672 & \bnum{1.24740} & \bnum{9.19271} \\
 
 Image variance \cite{Gallego17ral} & 1.69705397 & \bnum{11.252745982697} & \bnum{2.1798} & \bnum{21.9064} & \bnum{1.927} & \bnum{15.835} & 1.8214767814 & 15.8914689 \\

 Avg. timestamp \cite{Zhu19cvpr} & {\textgreater{}} {99} & {\textgreater{}} {99} & {\textgreater{}} {99} & {\textgreater{}} {99} & {\textgreater{}} {99} & {\textgreater{}} {99} & {\textgreater{}} {99}  &  {\textgreater{}} {99} \\

 Norm. avg. timestamp \cite{Paredes21neurips} & {\textgreater{}} {99} & {\textgreater{}} {99} & {\textgreater{}} {99} & {\textgreater{}} {99}  & {\textgreater{}} {99} &  {\textgreater{}} {99} & {\textgreater{}} {99} &  {\textgreater{}} {99} \\

\bottomrule
\end{tabular}
}
\end{table}

%% file: floats/fig_suppl_loss.tex
\def\figWidth{0.19\linewidth}
\begin{figure*}[h!]
	\centering
    {\scriptsize
    \setlength{\tabcolsep}{1pt}
	\begin{tabular}{
	>{\centering\arraybackslash}m{\figWidth} 
 	>{\centering\arraybackslash}m{\figWidth}
 	>{\centering\arraybackslash}m{\figWidth}
	>{\centering\arraybackslash}m{\figWidth}
	>{\centering\arraybackslash}m{\figWidth}}
		\\%\addlinespace[1ex]

		\gframe{\includegraphics[width=\linewidth]{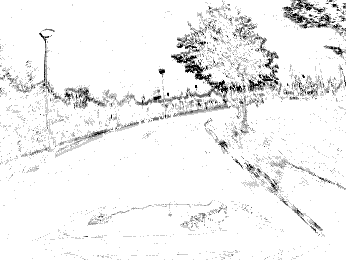}}
 		&\gframe{\includegraphics[width=\linewidth]{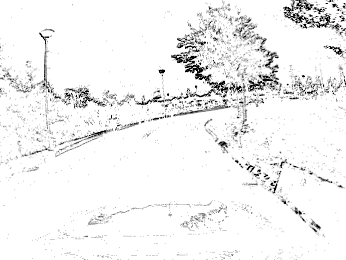}}
 		&\gframe{\includegraphics[width=\linewidth]{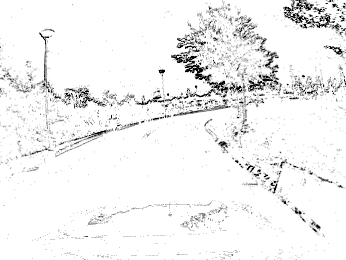}}
		&\gframe{\includegraphics[width=\linewidth]{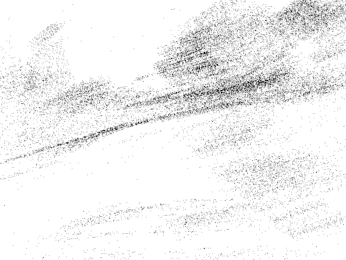}}
		&\gframe{\includegraphics[width=\linewidth]{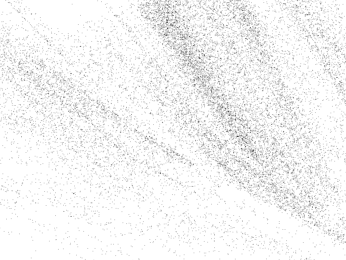}}
		\\

		(a) Input events
 		& (b) Gradient Magnitude
 		& (c) Variance
		& (d) Avg. timestamp \cite{Zhu19cvpr}
		& (e) Norm. avg. timestamp \cite{Paredes21neurips}
		\\%\addlinespace[1ex]
	\end{tabular}
	}
	\caption{
	\emph{IWEs for different loss functions}.
	Average timestamp losses overfit to undesired global optima, which pushes most events out of the image plane.
	}
\label{fig:suppl_loss}
\end{figure*}

%% file: floats/tab_suppl_reg_weight.tex
\begin{table}[h!]
\centering
\caption{Sensitivity analysis on the regularizer weight (MVSEC data, $dt=4$).
}
\label{tab:suppl_reg_weight}
\adjustbox{max width=\textwidth}{%
\setlength{\tabcolsep}{2pt}
\begin{tabular}{l*{8}{S[table-format=3.4]}}
\toprule
  & \multicolumn{2}{c}{indoor\_flying1} 
  & \multicolumn{2}{c}{indoor\_flying2} 
  & \multicolumn{2}{c}{indoor\_flying3} 
  & \multicolumn{2}{c}{outdoor\_day1}\\
 \cmidrule(l{1mm}r{1mm}){2-3}
 \cmidrule(l{1mm}r{1mm}){4-5}
 \cmidrule(l{1mm}r{1mm}){6-7}
 \cmidrule(l{1mm}r{1mm}){8-9}
 &\text{AEE $\downarrow$} & \text{\%Out $\downarrow$} & \text{AEE $\downarrow$} & \text{\%Out $\downarrow$}  
 &\text{AEE $\downarrow$} & \text{\%Out $\downarrow$} & \text{AEE $\downarrow$} & \text{\%Out $\downarrow$}  
 \\
\midrule

 \text{$\lambda=0.0025$} & 1.68217 & 12.79074 & 2.48917 & 26.31331 & 2.05633 & 18.92672 &  1.24740 &  9.19271 \\

 \text{$\lambda=0.025$} & 1.517299916933 & 9.07235138 & 2.3878414 & 26.25975 & 1.935548 &  18.440944647 & 1.85784597 & 17.114393846 \\
 
 \text{$\lambda=0.25$} & 1.885701460289 & 16.538685522128567 & 3.18600287 & 36.948295 & 2.9070675182 & 30.848186374 &  2.569955987360 & 27.86256470792523 \\

\bottomrule
\end{tabular}
}
\end{table}

%% file: floats/tab_suppl_dsec.tex
\begin{table}[h]
\vspace{-2ex} %
\begin{center}
\caption{Results on the DSEC optical flow benchmark \cite{Gehrig21threedv}.}
\label{tab:suppl_dsec}
\adjustbox{max width=1.0\textwidth}{%
\setlength{\tabcolsep}{3pt}
\begin{tabular}{l*{12}{S[table-format=2.2]}}
\toprule
  & \multicolumn{3}{c}{All}
  & \multicolumn{3}{c}{interlaken\_00\_b}
  & \multicolumn{3}{c}{interlaken\_01\_a}
  & \multicolumn{3}{c}{thun\_01\_a} \\
 \cmidrule(l{1mm}r{1mm}){2-4}
 \cmidrule(l{1mm}r{1mm}){5-7}
 \cmidrule(l{1mm}r{1mm}){8-10}
 \cmidrule(l{1mm}r{1mm}){11-13}
 
 & \text{AEE $\downarrow$} & \text{\%Out $\downarrow$} & \text{FWL $\uparrow$}
 & \text{AEE $\downarrow$} & \text{\%Out $\downarrow$} & \text{FWL $\uparrow$}
 & \text{AEE $\downarrow$} & \text{\%Out $\downarrow$} & \text{FWL $\uparrow$}
 & \text{AEE $\downarrow$} & \text{\%Out $\downarrow$} & \text{FWL $\uparrow$} \\
\midrule

 E-RAFT \cite{Gehrig21threedv}
 & 0.788 & 2.684 & 1.286231422619447
 & 1.394 &6.189 & 1.3233361693404235
 & 0.899 & 3.907 & 1.4233908392179435
 & 0.654 & 1.87 & 1.200584411 \\ 
 Ours
 & 3.472 & 30.855 & \bnum{1.365150407655252}
 & 5.744 & 38.925 & \bnum{1.4992904511327119}
 & 3.743 & 31.366 & \bnum{1.5137225860194}
 & 2.116 & 17.684 & \bnum{1.2420968732859203} \\ 

 \\[-0.2ex]

\midrule
  & \multicolumn{3}{c}{thun\_01\_b}
  & \multicolumn{3}{c}{zurich\_city\_12\_a}
  & \multicolumn{3}{c}{zurich\_city\_14\_c}
  & \multicolumn{3}{c}{zurich\_city\_15\_a} \\

 \cmidrule(l{1mm}r{1mm}){2-4}
 \cmidrule(l{1mm}r{1mm}){5-7}
 \cmidrule(l{1mm}r{1mm}){8-10}
 \cmidrule(l{1mm}r{1mm}){11-13}

 & \text{AEE $\downarrow$} & \text{\%Out $\downarrow$} & \text{FWL $\uparrow$}
 & \text{AEE $\downarrow$} & \text{\%Out $\downarrow$} & \text{FWL $\uparrow$}
 & \text{AEE $\downarrow$} & \text{\%Out $\downarrow$} & \text{FWL $\uparrow$}
 & \text{AEE $\downarrow$} & \text{\%Out $\downarrow$} & \text{FWL $\uparrow$} \\
\midrule

 E-RAFT \cite{Gehrig21threedv}
 & 0.578 & 1.518 & 1.1767931182449056
 & 0.612 & 1.057 & 1.1161122798926002
 & 0.713 &1.913 & 1.4688118
 & 0.589 & 1.303 & 1.335840906676725 \\ 
 Ours
 & 2.48 & 23.564 & \bnum{1.241943689019936}
 & 3.862 & 43.961 & \bnum{1.1375111350019202}
 & 2.724 & 30.53 & \bnum{1.4985924122973489}
 & 2.347 & 20.987 & \bnum{1.4119389239425226} \\

\bottomrule
\end{tabular}
}
\vspace{-8ex} %
\end{center}
\end{table}

%% file: floats/fig_suppl_extra_quality.tex
\def\figWidth{0.24\linewidth}
\vspace{-3ex} %
 \begin{figure*}[h!]
	\centering
    {\scriptsize
    \setlength{\tabcolsep}{1pt}
	\begin{tabular}{
	>{\centering\arraybackslash}m{\figWidth} 
	>{\centering\arraybackslash}m{\figWidth}
	>{\centering\arraybackslash}m{\figWidth}
	>{\centering\arraybackslash}m{\figWidth}}
		\\%\addlinespace[1ex]

		\gframe{\includegraphics[trim={0 66px 0 0},clip,width=\linewidth]{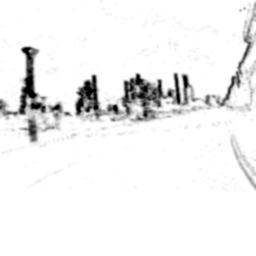}}
		&\gframe{\includegraphics[trim={0 66px 0 0},clip,width=\linewidth]{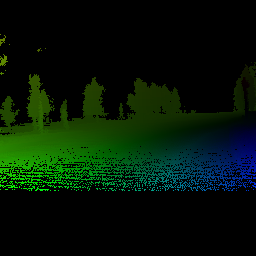}}
		&\gframe{\includegraphics[trim={0 66px 0 0},clip,width=\linewidth]{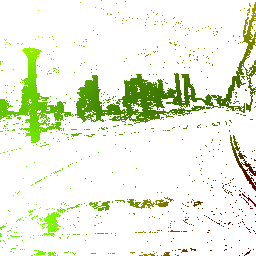}}
		&\gframe{\includegraphics[trim={0 66px 0 0},clip,width=\linewidth]{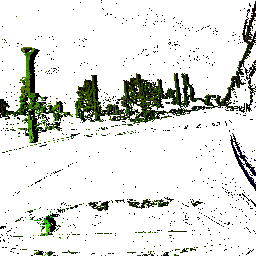}}		\\

		\gframe{\includegraphics[trim={0 66px 0 0},clip,width=\linewidth]{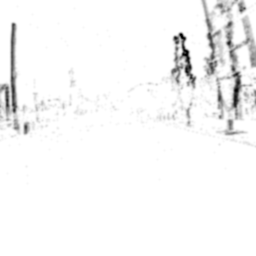}}
		&\gframe{\includegraphics[trim={0 66px 0 0},clip,width=\linewidth]{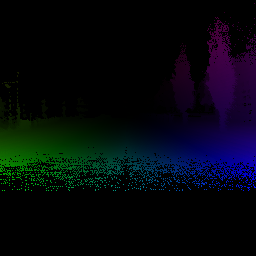}}
		&\gframe{\includegraphics[trim={0 66px 0 0},clip,width=\linewidth]{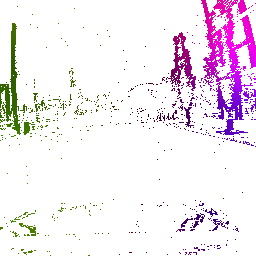}}
		&\gframe{\includegraphics[trim={0 66px 0 0},clip,width=\linewidth]{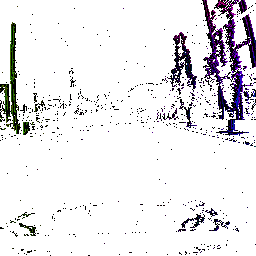}}
		\\

		(a) Input events
		& (b) GT
		& (c) Our EVFlowNet with \eqref{eq:composite:function}
		& (d) USL\cite{Paredes21neurips}
		\\%\addlinespace[1ex]
	\end{tabular}
	}
	\caption{\emph{Result of our DNN on the MVSEC outdoor sequence}.
	Our DNN (EV-FlowNet architecture) trained with \eqref{eq:composite:function} produces better result than the state-of-the-art unsupervised learning method \cite{Paredes21neurips}.
	For a quantitative comparison, see Table \ref{tab:dnn}.
	}
\label{fig:suppl_extra_qualitative}
\end{figure*}